\documentclass[letterpaper, 10 pt, conference]{ieeeconf}
\IEEEoverridecommandlockouts

\usepackage{bm}
\usepackage{acro}

\usepackage{xspace}

\usepackage{array}
\usepackage{booktabs}
\usepackage{fontawesome}
\usepackage{xcolor,colortbl}
\definecolor{Gray}{gray}{0.9}
\definecolor{Textgray}{gray}{0.4}

\definecolor{notetext}{rgb}{0.7,0,0}
\definecolor{notetext}{rgb}{0.7,0,0}

\usepackage[nointegrals]{wasysym}
\usepackage{makecell}
\usepackage{balance}
\usepackage{multirow}
\usepackage{overpic}
\usepackage{bbm}
\usepackage{dsfont}
\usepackage[font=small,labelfont=bf]{caption}
\usepackage{color}
\usepackage{pifont}
\usepackage{etoolbox}
\usepackage{float}
\usepackage{tikz}
\usepackage{svg}
\usepackage{url}
\usepackage{times}
\usepackage{epsfig}
\usepackage{graphicx}
\usepackage{amsmath}
\usepackage{amssymb}
\usepackage{comment}
\usepackage{soul}
\usepackage{listings}
\usepackage{lscape}
\usepackage{rotating}
\let\labelindent\relax
\usepackage{enumitem}
\usepackage[utf8]{inputenc}
\usepackage{pgfplots}
\usepackage{import}
\usepackage{xifthen}
\usepackage{transparent}
\usepackage{tabularx}
\usepackage{tabu}
\usepackage{xcolor}
\usepackage{arydshln}
\DeclareUnicodeCharacter{2212}{−}
\usepgfplotslibrary{groupplots,dateplot}
\usetikzlibrary{patterns,shapes.arrows}
\pgfplotsset{compat=newest}
\usepackage{amssymb}
\usepackage{pifont}
\newcommand{\cmark}{\ding{51}}%

\newcommand{\xmark}{\ding{55}}%
\usepackage{silence} 
\usepackage{subcaption}
\usepackage{xfrac}
\usepackage{tcolorbox}

\definecolor{m_red}{RGB}{255,209,209}
\definecolor{m_red_border}{RGB}{215,23,20}

\definecolor{m_orange}{RGB}{226,213,231}
\definecolor{m_orange_border}{RGB}{150,114,164}

\definecolor{m_blue}{RGB}{217,232,251}
\definecolor{m_blue_border}{RGB}{107,141,190}

\definecolor{m_yellow}{RGB}{255,242,205}
\definecolor{m_yellow_border}{RGB}{213,182,82}

\definecolor{m_gray}{RGB}{245,245,245}
\definecolor{m_gray_border}{RGB}{102,102,102}

\lstdefinestyle{mypython}{
  language=python,
  breaklines=true,
  basicstyle=\fontsize{8}{12}\selectfont\ttfamily,
  keywordstyle=\bfseries\color{my_blue},
  linewidth=.99\textwidth,
}

\newcommand{\PAR}[1]{\vskip4pt \noindent {\bf #1~}}

\definecolor{darkgreen}{RGB}{0,255,0}
\definecolor{linkgreen}{RGB}{52,130,48}

\newcolumntype{Y}{>{\centering\arraybackslash}X}
\newcolumntype{Z}{>{\raggedleft\arraybackslash}X}

\newif\ifmynotes
\mynotestrue
\definecolor{notetext}{rgb}{0.7,0,0}

\makeatletter
\def\adl@drawiv#1#2#3{%
        \hskip.5\tabcolsep
        \xleaders#3{#2.5\@tempdimb #1{1}#2.5\@tempdimb}%
                #2\z@ plus1fil minus1fil\relax
        \hskip.5\tabcolsep}
\newcommand{\cdashlinelr}[1]{%
  \noalign{\vskip\aboverulesep
           \global\let\@dashdrawstore\adl@draw
           \global\let\adl@draw\adl@drawiv}
  \cdashline{#1}
  \noalign{\global\let\adl@draw\@dashdrawstore
           \vskip\belowrulesep}}
\makeatother

\makeatletter
\newcommand\smaller{\@setfontsize\smaller{8.7}{9.5}}
\makeatother

\title{Panoptic-CUDAL: Rural Australia Point Cloud Dataset in Rainy Conditions\\

\author{
{Tzu-Yun Tseng$^{1\dagger}$, Alexey Nekrasov$^{2\dagger}$, Malcolm Burdorf$^{2\dagger}$}
\\
{Bastian Leibe$^2$, Julie Stephany Berrio$^1$, Mao Shan$^1$, Zhenxing Ming$^1$, Stewart Worrall$^1$}
}

\thanks{\textdagger ~ equal contribution}
\thanks{$^1$ The Australian Centre for Robotics (ACFR) at the University of Sydney (NSW, Australia)}
\thanks{$^2$ RWTH Aachen University (Aachen, Germany)}
\thanks{This research was conducted as part of the project titled ``Data analytics tools for developing and testing autonomous urban road vehicle capabilities" and was partially supported by the Australian Government through the Australian Research Council's ARC Training Centre funding scheme for Automated Vehicles in Rural and Remote Regions  (project IC230100001)}
}

\begin{document}
\maketitle
\thispagestyle{empty}
\pagestyle{empty}

\begin{abstract}
%
Existing autonomous driving datasets are predominantly oriented towards well-structured urban settings and favourable weather conditions, leaving the complexities of rural environments and adverse weather conditions largely unaddressed.
Although some datasets encompass variations in weather and lighting, bad weather scenarios do not appear often.
Rainfall can significantly impair sensor functionality, introducing noise and reflections in LiDAR and camera data and reducing the system's capabilities for reliable environmental perception and safe navigation.
This paper introduces the Panoptic-CUDAL dataset, a novel dataset purpose-built for panoptic segmentation in rural areas subject to rain.
By recording high-resolution LiDAR, camera, and pose data, Panoptic-CUDAL offers a diverse, information-rich dataset in a challenging scenario.
We present the analysis of the recorded data and provide baseline results for panoptic, semantic segmentation, and 3D occupancy prediction methods on LiDAR point clouds.
The dataset can be found here: 
\url{https://vision.rwth-aachen.de/panoptic-cudal}
\end{abstract}



\section{Introduction}
Autonomous driving has received significant attention in recent years, with multiple datasets~\cite{cordts2016cityscapes,neuhold2017mapillary,behley2019semantickitti,caesar2020nuscenes,sun2020waymo} and methods~\cite{marcuzzi2023maskpls,yilmaz2024mask4former,wu2022ptv2,xu2025frnet,zhou2020cylinder3d,cortinhal2020salsanext}.
Although high-quality data has contributed significantly to model improvement~\cite{sun2020waymo}, there is still a need for challenging training and evaluation data in real-world conditions.
We aim to address the need for modern automotive datasets in different environments and under challenging weather conditions.

\begin{figure}[t!]
\centering
\includegraphics[width=0.99\columnwidth]{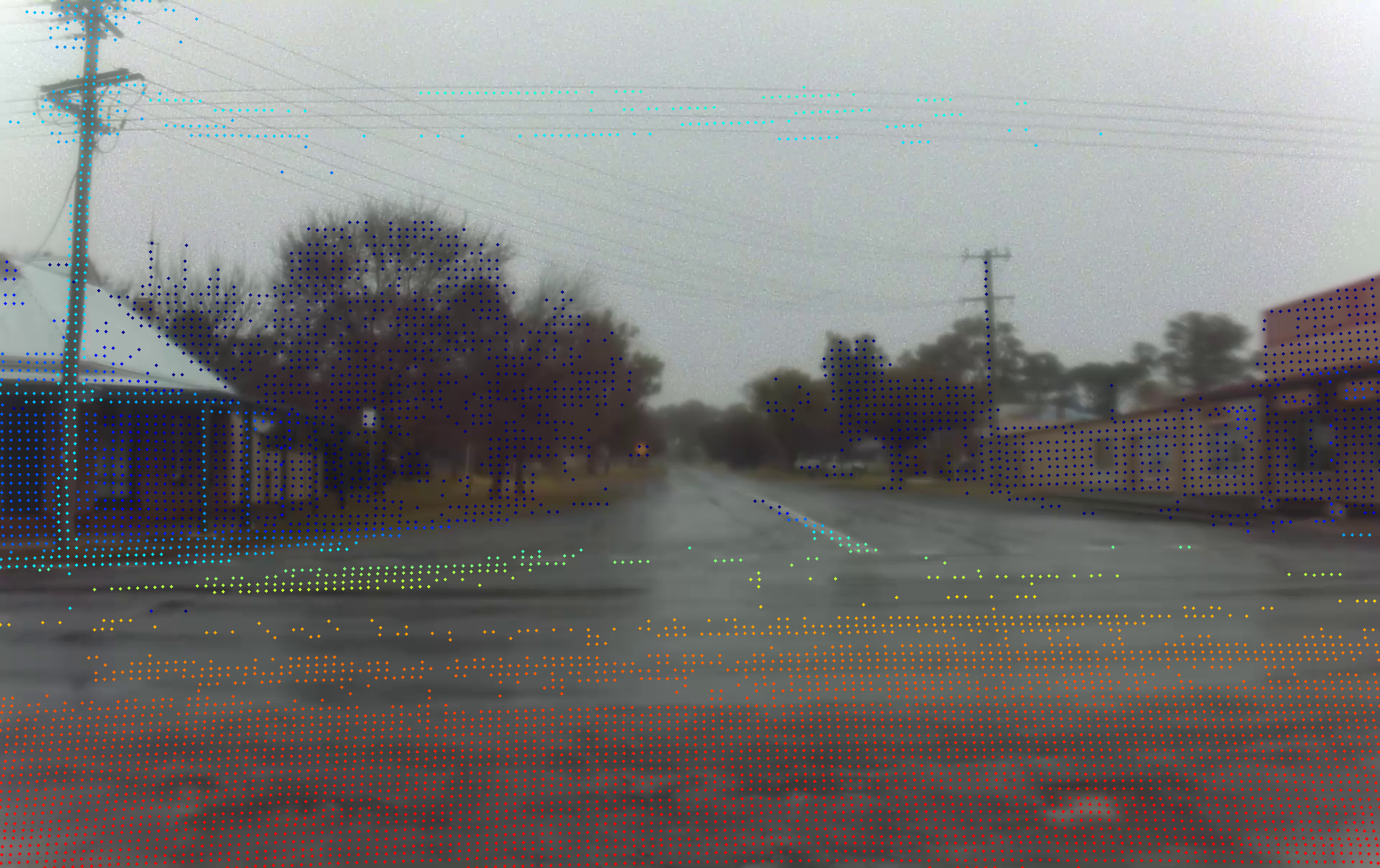}
\caption{
Front-view image from the Panoptic-CUDAL dataset, with projected LiDAR point clouds.
}
\label{fig:teaser}
\vspace{-4mm}
\end{figure}

Data collection and annotation face significant limitations due to high costs, particularly for datasets consisting of various data distributions~\cite{neuhold2017mapillary}.
Although urban driving scenarios are well represented in existing datasets, remote environments remain understudied despite the important role of rural roads in transportation networks.
Recent community interest in thoroughly annotated rural driving datasets~\cite{ninan2023r2d2,almin2023navya3dseg} highlights the need for more research in this direction.
Especially given the unique challenges posed by specific characteristics of non-urban settings.


Rural and urban environments pose different challenges to autonomous vehicles.
Urban datasets introduce multiple instance of objects in the same scene, multiple classes of long tails, and complex interaction of objects on a large scale~\cite{sun2020waymo}.
However, in contrast to urban spaces, rural datasets introduce other challenges, such as vast open spaces, different road infrastructure, and sometimes reduced levels of road maintenance.
This difference becomes particularly pronounced in adverse weather conditions, where model performance typically degrades significantly~\cite{sakaridis2024acdc}.
This includes rain, which can create large puddles on rural roads, snow, or fog.
Addressing these challenges requires comprehensive datasets that capture the long-tail distribution of real-world driving scenarios while enabling model evaluation.


To address the need for a challenging dataset, we present Panoptic-CUDAL, a new LiDAR dataset captured in a rural area during rain.
We record six driving sequences in Cudal, a town in the state of New South Wales in Australia (seen in Figure~\ref{fig:teaser}).
Our dataset features synchronized sensor data, including surround-view camera images, high-resolution LiDAR point clouds, and odometry information.
The data collected has diverse, challenging scenarios inherent to rural environments.
We annotate the dataset with panoptic labels following the SemanticKITTI~\cite{behley2019semantickitti} labels format.


\begin{table*}[ht]
\centering
    \caption{
    Comparison of Automotive Datasets with Panoptic Labels on the LiDAR Level. 
    Existing works focus on urban scenes with no rainfall.
    }

    \begin{tabularx}{1.0\linewidth}{l c YYYY cc}
    \toprule
    \textbf{Dataset} & \textbf{Year} & \textbf{LiDAR Beams} & \textbf{LiDAR Annotations} & \textbf{Annotated Scans} & \textbf{Surround-View Cameras} & \textbf{Rural} & \textbf{Rain}  \\
    \midrule
    SemanticKITTI~\cite{behley2019semantickitti} & 2019 & 64 & Panoptic & 15k      & \xmark & \xmark & \xmark                 \\
    Waymo~\cite{sun2020waymo} & 2019 & 64 & Semantic & 230k    & \cmark & \xmark & \cmark \\
    Panoptic nuScenes~\cite{fong2021panoptic} & 2021 & 32 & Panoptic & 40k     & \cmark & \xmark & \xmark    \\
    Navya3DSeg~\cite{almin2023navya3dseg} & 2023 & 64 & Semantic & 50k & \xmark & \cmark & \xmark \\
    R2D2~\cite{ninan2023r2d2} & 2023 & 128 & Semantic & 10.5k   & \xmark & \cmark & \xmark \\
    Panoptic-CUDAL (ours) & 2025 & 128 & Panoptic & 14.7k   & \cmark & \cmark & \cmark \\
    \bottomrule
    \end{tabularx}
\end{table*}

The combination of high-resolution LiDAR (128-channel), surround-view camera images, large dataset size, and data intentionally collected under rainfall conditions in a rural area addresses the need for a dataset to train and validate models operating in these challenging environments.
Accurately annotated data allow the evaluation of semantic segmentation~\cite{xu2025frnet,milioto2019rangenet,wu2022ptv2}, panoptic~\cite{yilmaz2024mask4former,marcuzzi2023maskpls} segmentation, and 3D occupancy prediction ~\cite{zhang2024occfusion}, ~\cite{ming2024inversematrixvt3d} methods for static and dynamic objects.
We analyze the dataset labels, qualitatively show challenging scenarios, train, and evaluate a set of models for semantic segmentation, panoptic segmentation and 3D occupancy prediction on the proposed dataset.
Our models achieve acceptable performance on the proposed dataset, and we expect that the recorded data will be used by the community to train and evaluate segmentation methods to address challenges posed by rural roads and adverse weather conditions.

To summarize, the main contributions of this paper are as follows:
\begin{itemize}
  \item We introduce Panoptic-CUDAL, the first rural autonomous driving dataset with panoptic labels captured in rainy conditions, integrating a high-resolution 128-beam LiDAR, 8 surround-view cameras, and pose data.
  \item The dataset is released in the SemanticKITTI format ~\cite{behley2019semantickitti}, ensuring compatibility with standard tools and benchmarks.
  \item We benchmark multiple state-of-the-art models for semantic segmentation, panoptic segmentation, and 3D occupancy prediction on this dataset.
  \item A detailed performance analysis across semantic classes, highlighting current model limitations in rare-class scenarios and under adverse weather conditions.
\end{itemize}

\section{Related Work}

\textbf{Automotive Datasets.}
The autonomous driving community has witnessed substantial progress through the development of datasets over recent years. 
Early efforts focused primarily on camera-based solutions, often employing a single front-facing monocular~\cite{brostow2009camvid,leibe2007leuven} or stereo vision system~\cite{cordts2016cityscapes}, with a low-resolution LiDAR sensor~\cite{maddern2017oxfordcar}.
Further works introduced multimodal sensor setups~\cite{geiger2013kitti,caesar2020nuscenes,behley2019semantickitti,mao2021once} incorporating high-resolution LiDAR, radars, and surround-view camera setups.
Parallel advancements have emerged in geographical diversity, with datasets expanding coverage to global regions and with dense labeling schemes~\cite{varma2019idd,neuhold2017mapillary,yu2020bdd100k}.
In particular, most existing works do not explicitly identify the weather conditions or complexity of a scenery in their recordings.

\textbf{Adverse Datasets.}
Recent years have seen increased focus on adverse weather perception, with image-based datasets addressing various visual conditions such as nighttime~\cite{dai2018nightdriving,sakaridis2022nighttime}, hazards environments~\cite{zendel2018wilddash}, or focus on seasonal changes~\cite{sakaridis2024acdc,wenzel20204seasons}. 
However, rural environment introduce challenges, such as ambiguous road boundaries and sparse infrastructure, which remain understudied in the autonomous driving context.  
The ORFD dataset~\cite{min2022orfd} partially addresses this gap but limits its scope only to drivable area segmentation.  
While RUGD~\cite{wigness2019rugd} provides multimodal rural data, it is collected using a small robotics platform that makes it unsuitable for full-size vehicle perception.  
Recent efforts like Navya3DSeg~\cite{almin2023navya3dseg} explicitly include limited rural and residential areas in their data collection. 
The R2D2~\cite{ninan2023r2d2} dataset specifically targets rural road challenges through dense LiDAR semantic annotations.  
However, recent works lack comprehensive coverage of scenarios in rural contexts under rain conditions, and do not provide panoptic annotations or extensive sensor setup.

\textbf{LiDAR Segmentation Methods.}
Due to the challenging nature of point cloud data, several segmentation approaches have emerged in recent years to segment LiDAR point clouds using different underlining representations.
Recent LiDAR segmentation methods have attempted to solve the task using projection-based representations~\cite{milioto2019rangenet,cortinhal2020salsanext}, exploring voxelized point cloud representation~\cite{choy2019minkowski,zhou2020cylinder3d}, explore other efficient representations~\cite{wu2022ptv2,wu2024ptv3} and their combinations~\cite{xu2025frnet,hou2022pointtovoxel}, or explore combinations with different modalities~\cite{yan20222dpass}.
To complement the semantic task, the task of panoptic segmentation requires segmentation of individual object instances~\cite{fong2021panoptic} in addition to semantic classes.
In the field of outdoor LiDAR point cloud segmentation, several panoptic segmentation works were proposed~\cite{yilmaz2024mask4former,marcuzzi2023maskpls,schult2023mask3d,sirohi2022efficientlps,fong2021panoptic,su2023pups}.
With recent transformer-based methods~\cite{marcuzzi2023maskpls,yilmaz2024mask4former,schult2023mask3d}.
Our work evaluates semantic segmentation, panoptic segmentation, and 3D occupancy prediction, establishing baseline performance on the proposed dataset.

\section{Dataset}

\subsection{Data Collection Vehicle.}
Our data collection platform runs ROS \cite{ros} and stores all data in rosbag files.
We primarily use two types of sensors positioned on the vehicle: cameras and LiDAR.

\begin{figure*}[t]
    \centering
	\includegraphics[trim={0 0 0 1.3cm},clip, width=0.95\textwidth]{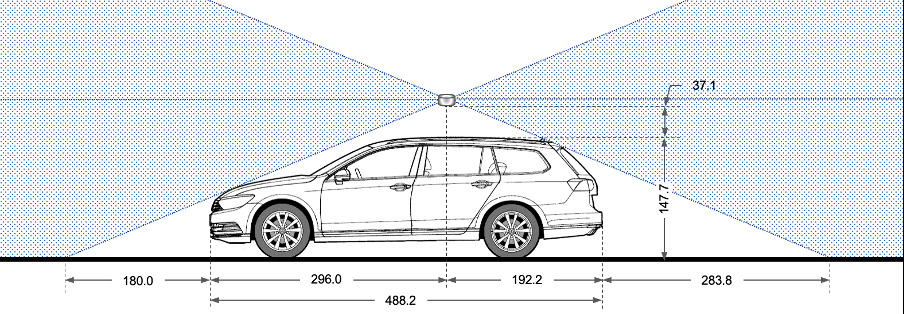}
    \caption{
        The LiDAR is located on the vehicle's roof used to collect the data.
    }
    \label{fig:ibeo_ous}
\end{figure*}

\textbf{LiDAR.}
An Ouster OS1-128 LiDAR sensor is placed on the vehicle roof, ensuring a full 360$^{\circ}$ view of the environment.
The OS1-128 provides $128$ vertical beams within a $45^{\circ}$ vertical field of view (FOV), with the sampling rate of 10 Hz, that can detect objects up to $200$ meters away, ensuring high resolution of captured point clouds.
Figure \ref{fig:ibeo_ous} illustrates the placement of this 3D LiDAR on the vehicle.

\begin{figure*}[t]
    \centering
    \includegraphics[width=\linewidth]{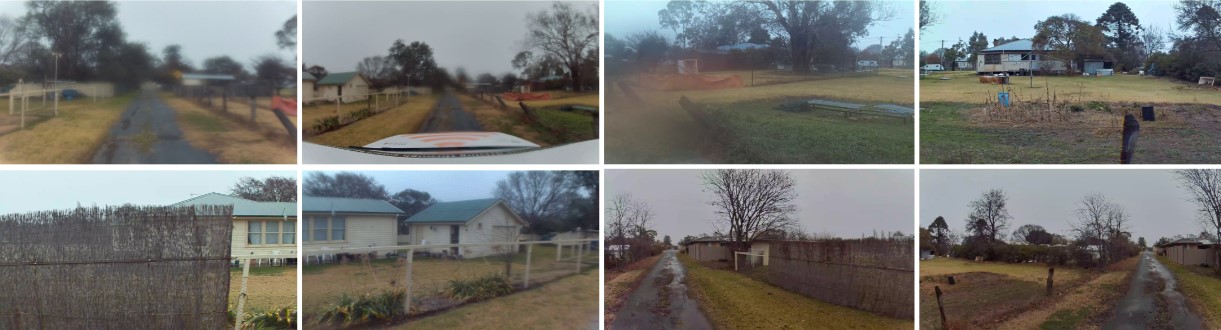}
    \caption{
    \textbf{Surround-view images}
    Our vehicle is equipped with 8 cameras ensuring 360$^{\circ}$ view of the environment.
    }
    \label{fig:cameras360}
\end{figure*}
\begin{figure}[t]
    \centering
    \includegraphics[width=0.9\columnwidth]{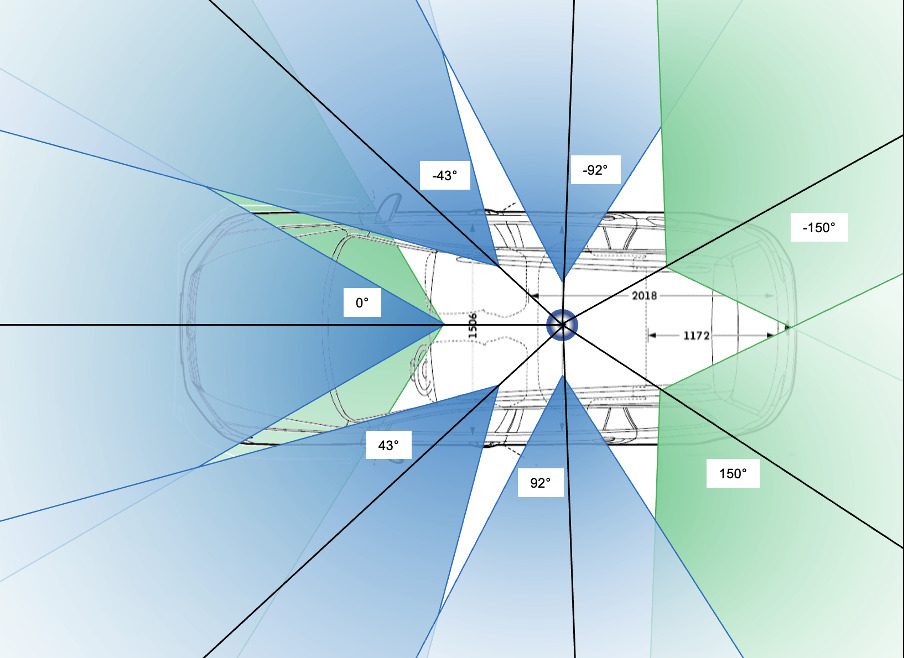}
    \caption{
    \textbf{Camera configuration}
    Our vehicle is equipped with cameras ensuring 360$^{\circ}$ view of the environment.
    The blue and green colour represents the 60$^{\circ}$ and 120$^{\circ}$ FOV of the cameras, respectively.
    }
    \label{fig:cameras3}
    \vspace{-4mm}
\end{figure}

\textbf{Cameras.}
Eight cameras, consisting of three SF3324 and five SF3325 automotive GMSL units, are mounted on the collection platform.
Each camera features an ONSEMI AR0231 CMOS image sensor (2.3M pixels) paired with a SEKONIX ultra-high-resolution lens.
The lenses have a horizontal field of view (FOV) of 60$^{\circ}$ or 120$^{\circ}$ and a vertical FOV of 38$^{\circ}$ or 73$^{\circ}$.
The images were recorded at 30 frames per second (fps) with a resolution of 1928$\times$1208 pixels.

The cameras are arranged horizontally, allowing 360$^{\circ}$ coverage of the vehicle’s surroundings (shown on Figure~\ref{fig:cameras360} and Figure~\ref{fig:cameras3}).
Two front-center cameras with a narrow and wide field of view ensure perception at short and long distances.
Front-center cameras are oriented parallel to the ground, such that the lowest visible points appear approximately $1.8$ meters away from the vehicle.
The front left and front right cameras angled at the sides with narrow fields of view improve side coverage at the front of the vehicle.
Side cameras, mounted on the left and right sides of the vehicle provide lateral views.
Two wide-view rear-facing cameras complete the full 360$^{\circ}$ environmental coverage, ensuring that the vehicle perception system captures all necessary visual information.
To perform extrinsic calibration, we used our camera calibration method~\cite{surabhi_calib_2019} to determine the transformation matrices between each camera and the LiDAR coordinate frame.

\subsection{Location.}
We collected the data in Cudal, a rural town in Australia.
The data collected in Cudal differs from publicly available data, which typically focus on urban environments.
The town has a main street with smaller parallel and perpendicular side streets.
The vehicle was driven on a single rainy day to capture different daytime conditions and rain intensity.
To ensure coverage of the town, a single long sequence was recorded and later cut into individual sequences of fixed length.
Cudal's rural landscape is characterized by dense vegetation, sparsely distributed buildings, and roads that may be unpaved or lack clear lane markings.
These conditions present significant challenges for AV systems, as the driving surface is less defined.

\begin{figure}[t]
    \centering
    \begin{subfigure}[]{0.49\columnwidth}
    \centering
    \includegraphics[trim={0cm 0cm 0cm 0cm},clip,width=\columnwidth]{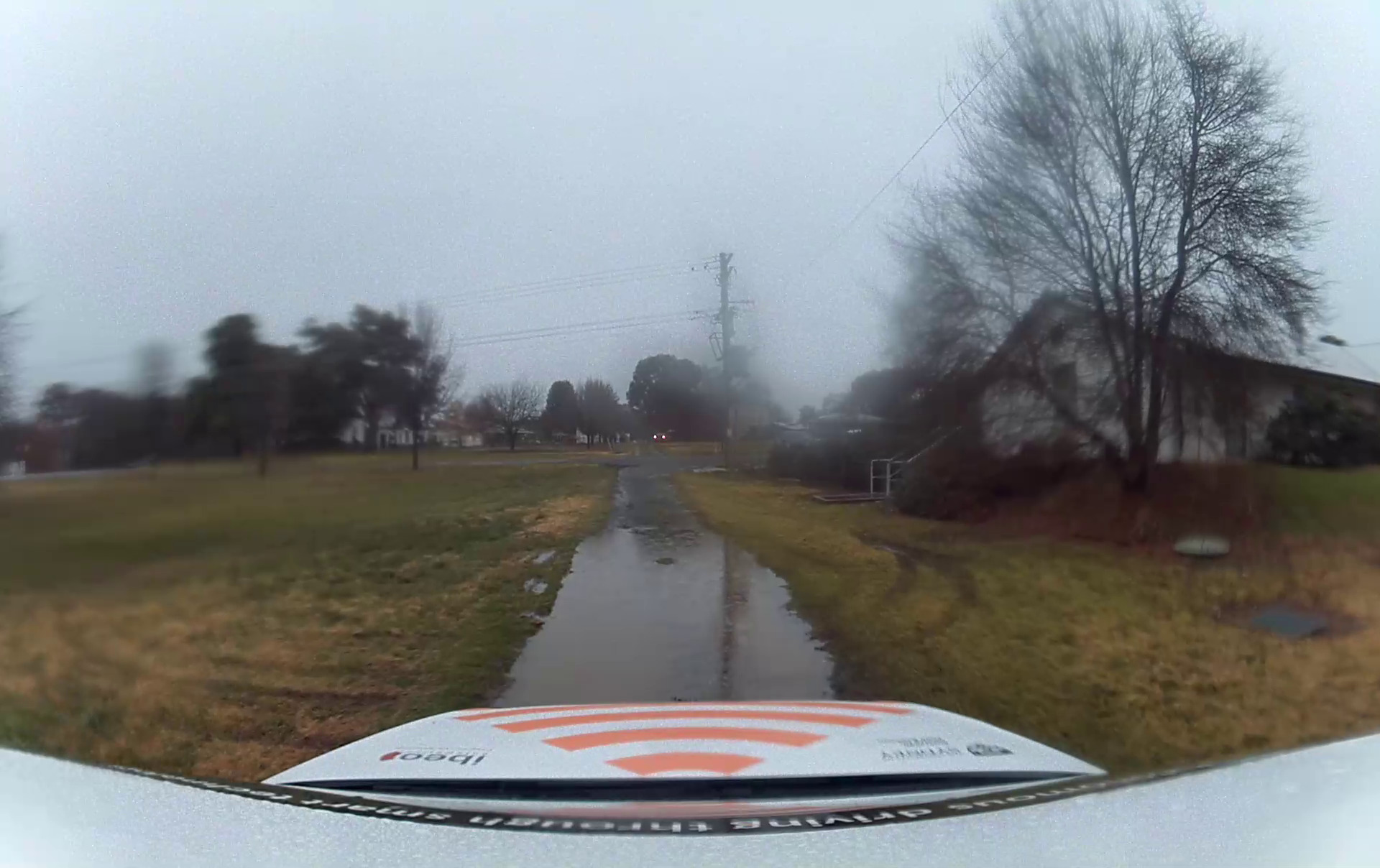}
    \caption{\small Puddle of water in front of the vehicle.}
    \label{fig:cudal_1}
    \end{subfigure}
    \begin{subfigure}[]{0.49\columnwidth}
    \centering
    \includegraphics[trim={0cm 0cm 0cm 0cm},clip,width=\columnwidth]{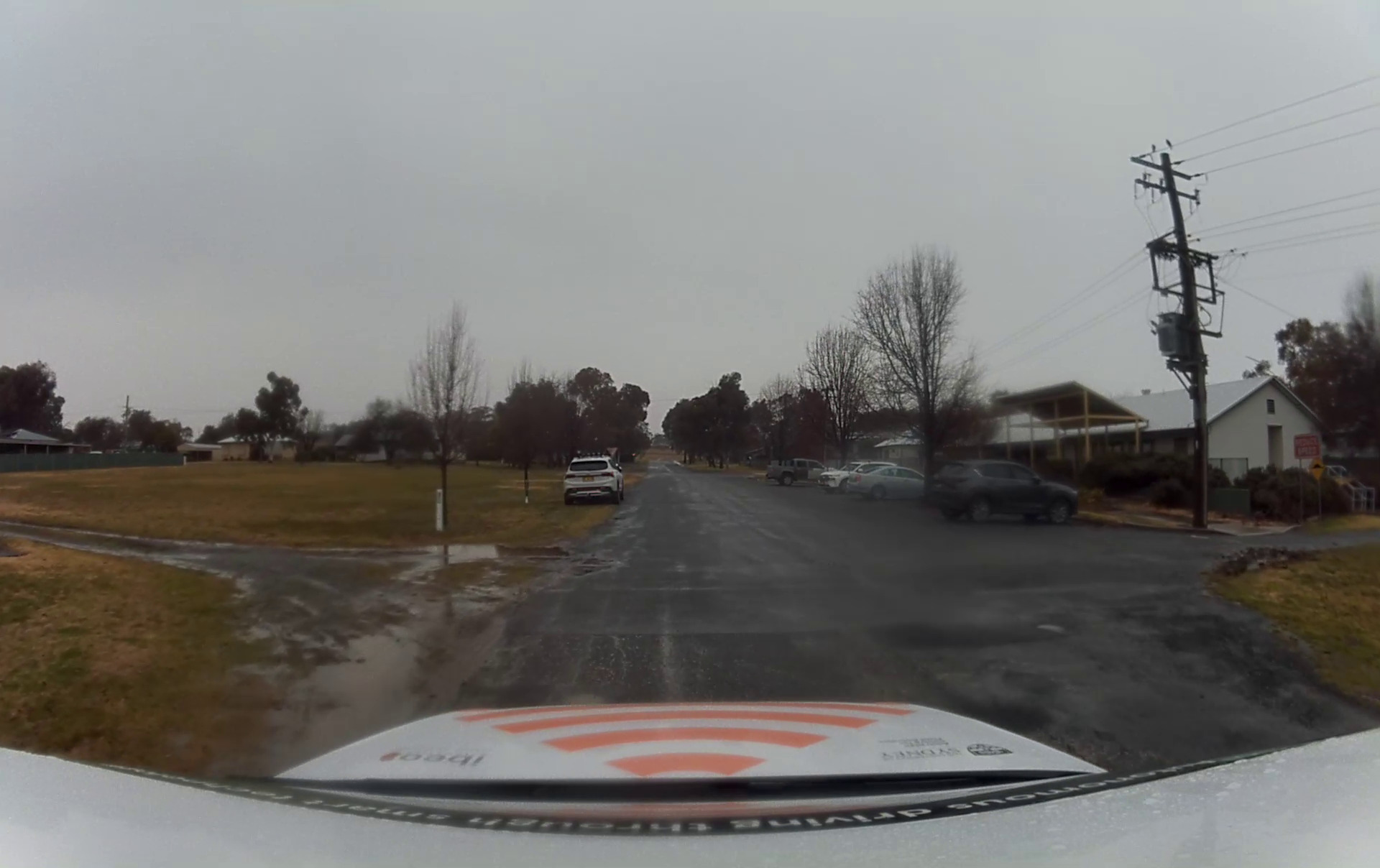}
    \caption{\small Road boundaries not well defined.}
    \label{fig:cudal_2}
    \end{subfigure}

    \vspace{5mm}

    \begin{subfigure}[]{0.49\columnwidth}
    \centering
    \includegraphics[width=\columnwidth]{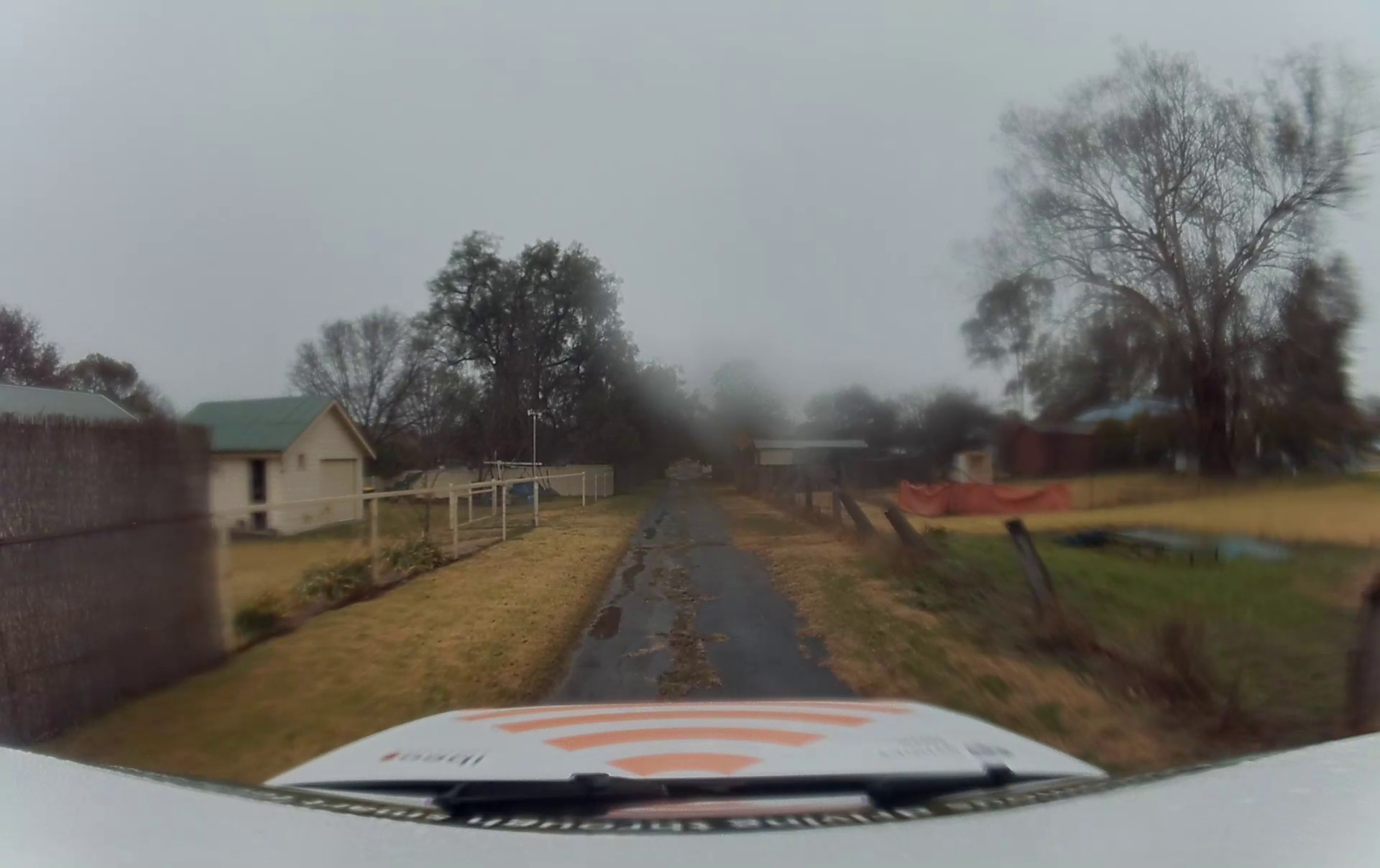}
    \caption{\small Narrow passage without lane markings.}
    \label{fig:cudal_3}
    \end{subfigure}
    \begin{subfigure}[]{0.49\columnwidth}
    \centering
    \includegraphics[width=\columnwidth]{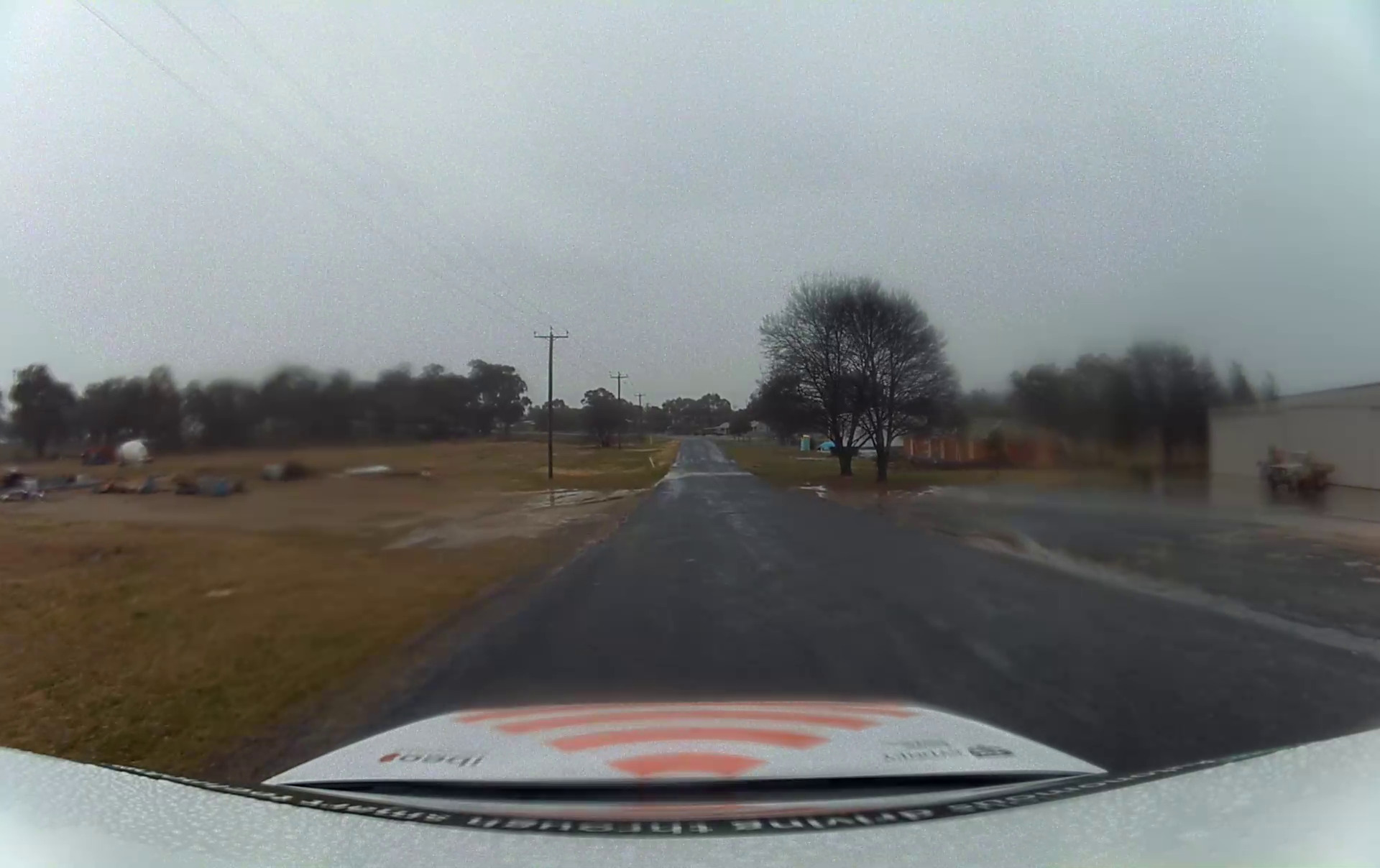}
    \caption{\small Wide open space in the driving environment.}
    \label{fig:cudal_4}
    \end{subfigure}
  
    \caption{\small 
        \textbf{Perception challenges for rural areas.}
        The recorded data contains multiple examples of challenging conditions, rarely observed in other datasets.
    }
    \label{fig:cudal_challenges}
\end{figure}

Figure~\ref{fig:cudal_challenges} highlights four challenges commonly encountered by autonomous vehicles in rural environments.
The absence of well-defined road boundaries, large areas with few reference objects, and the absence of lane markings pose difficulties to the correct segmentation of LiDAR point clouds.
Since the dataset is collected during rain, it additionally introduces corruptions to the camera frames.
This complicates driving in these conditions for humans and autonomous vehicles.

\begin{figure*}[t]
    \centering
    \includegraphics[width=\textwidth]{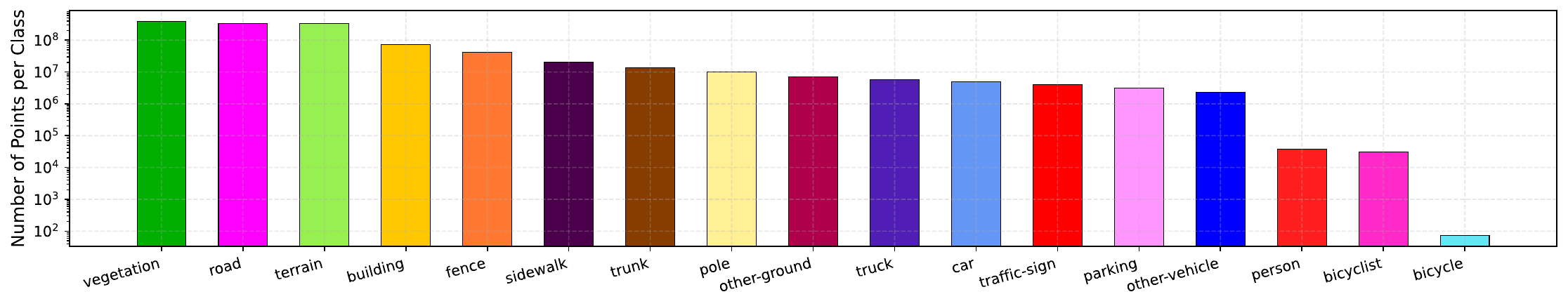}
    \caption{
    \textbf{Class distribution in the training set.}
    Due to the recording in a rural area, most of the points in the dataset belong to `stuff' classes.
    Only a few points assigned to `things', such as vehicles or other actors on roads.
    }
    \label{fig:labeldist}
\end{figure*}

\begin{figure}[t]
    \centering
    \includegraphics[width=\columnwidth]{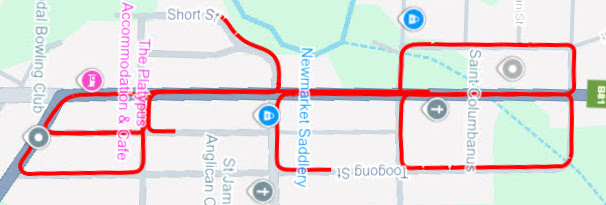}
    \caption{
        \small \textbf{Data collection routes.}
        Six sequences were recorded in the area of Cudal.
        }
    \label{fig:route}
\end{figure}

\subsection{Data Post-processing.}
\textbf{Vehicle localization.}
We used a prebuilt point cloud map generated by the LiDAR-Inertial Odometry via Smoothing and Mapping (LIO-SAM) method \cite{shan2020liosam} to achieve robust vehicle localisation in the Cudal environment.
The localisation process relied on a scan matching-based algorithm that aligns incoming LiDAR scans with this reference map to accurately determine the vehicle's position and orientation in real time.
Each new LiDAR scan is matched against the pre-existing high-resolution point cloud map using a scan-to-map registration strategy \cite{3d_ndt}, allowing precise estimation of the vehicle's position and orientation at each timestamp.
This scan-matching localisation method involved iterative optimisation techniques, where each real-time LiDAR scan was compared against the previously mapped environment to find the best-fit position and orientation.
By leveraging the spatial details captured by the prebuilt map, the localisation algorithm significantly reduced drift and enhanced accuracy, even under challenging environmental conditions.
We show recorded routes overlayed on the town map in Figure~\ref{fig:route}.

\textbf{LiDAR motion correction.}
Initially, all data points captured by the LiDAR sensor are represented within the LiDAR's own coordinate frame. However, because the LiDAR sensor and camera capture data at slightly different times, the raw LiDAR data must be corrected to ensure accurate alignment with the corresponding camera image.
To achieve this temporal alignment, we first determine the displacement (movement) of the vehicle that occurred between the LiDAR scan timestamp and the camera image timestamp. This displacement is then used to adjust the LiDAR individual points in the vehicle frame. This process synchronizes the point cloud to the exact moment when the camera captured the image \cite{berrio2020cameralidar}. 

After correcting for this temporal offset, we transform the adjusted point cloud back into the LiDAR's original coordinate frame. This final step maintains the spatial consistency of the data, ensuring that the LiDAR points accurately reflect object positions without distortions caused by vehicle motion.

\begin{figure}[t]
    \centering
    \includegraphics[width=0.99\columnwidth]{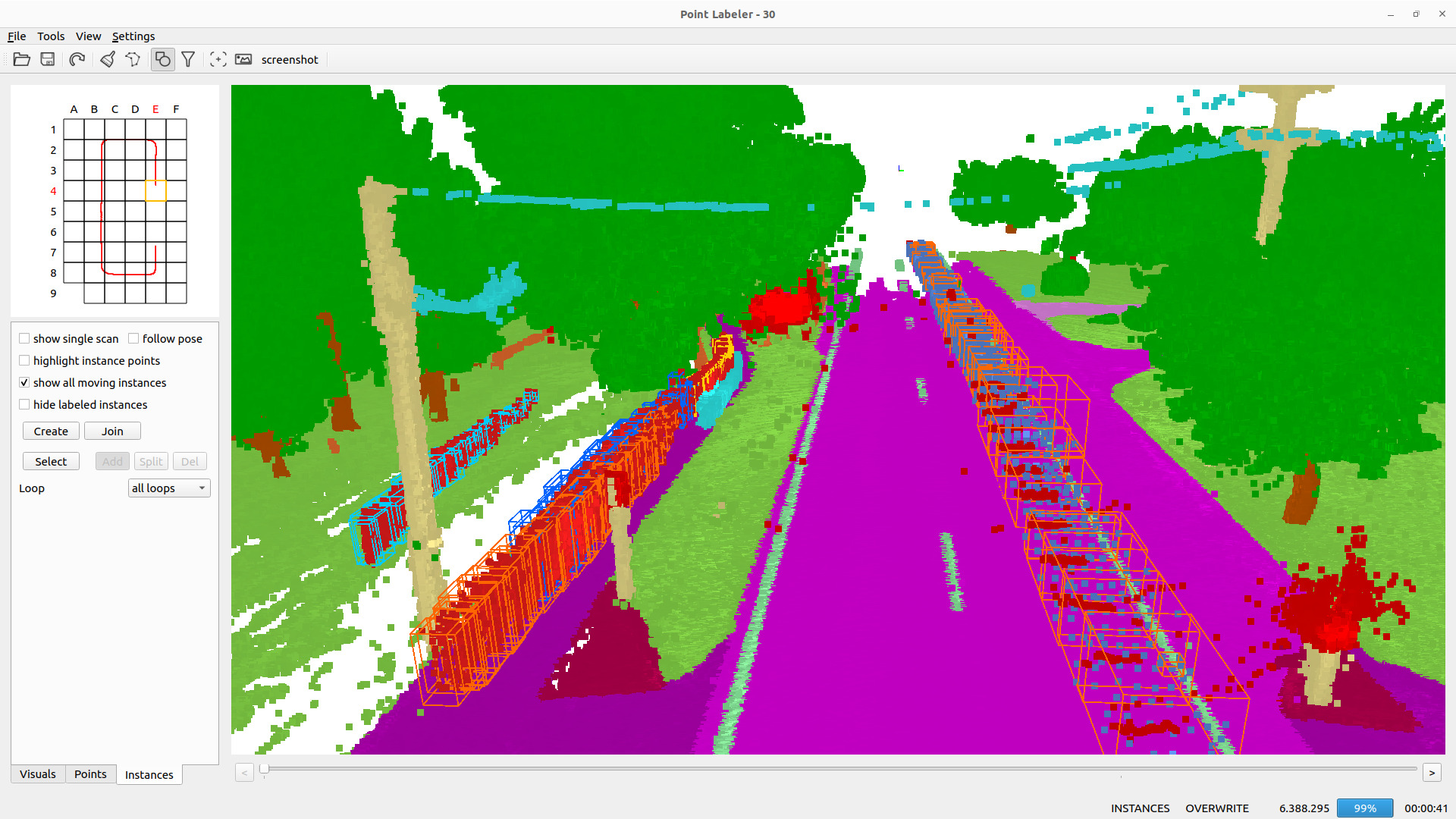}
    \caption{\small 
    Annotation tool~\cite{behley2019semantickitti} used to label the point cloud.
    We label point cloud with semantic classes per point and instance labels; shown in the labeling tool as boxes.
    }
    \label{fig:tool}
    \vspace{-4mm}
\end{figure}

\textbf{Formatting.}
We converted point clouds and ROSBag localization data to the SemanticKITTI~\cite{behley2019semantickitti} dataset format to take advantage of the annotation tools and benchmarks that the SemanticKITTI format supports.
This approach ensures compatibility with existing software tools and enables direct comparison with other research efforts.
The conversion process begins by extracting the relevant point cloud, images, and odometry topics from the ROSBag.
Point cloud data, generally stored as PCL (Point Cloud Library) structures \cite{rusu2011pcl}, is reorganized into the KITTI binary format.
This involves arranging the information into a structured array with fields for the x, y, z coordinates and intensity values.
Odometry data are extracted and converted into the SemanticKITTI pose format, which requires generating transformation matrices that represent the position and orientation of the vehicle over time.
As a result, each recorded sequence is stored in SemanticKITTI format, with LiDAR point clouds, ego-poses, and surround-view images for each timestamp.

\begin{table*}[t]
  \centering
  \resizebox{\textwidth}{!}{%
  \begin{tabularx}{1.0\linewidth}{lY c YYYYYYYYYYYYYYYYYYY}
    \toprule
    Method & 
    mIoU & 
    &
    \rotatebox{90}{Car} & 
    \rotatebox{90}{Truck} & 
    \rotatebox{90}{Other-Vhcl} & 
    \rotatebox{90}{Person} & 
    \rotatebox{90}{Road} &  
    \rotatebox{90}{Sidewalk} & 
    \rotatebox{90}{Other-Gnd} & 
    \rotatebox{90}{Building} & 
    \rotatebox{90}{Fence} & 
    \rotatebox{90}{Vegetation} &  
    \rotatebox{90}{Trunk} & 
    \rotatebox{90}{Terrain} & 
    \rotatebox{90}{Pole} & 
    \rotatebox{90}{Traffic-sign} \\
    \midrule
    Rangenet53++~\cite{milioto2019rangenet}           & 23.9 && 24.2 & 00.4 & 00.0 & 00.0 & 84.1 & 17.0 & 00.3 & 34.3 & 29.5 & 59.5 & 37.3 & 86.9 & 30.2 & 50.7 \\
    Rangenet21++~\cite{milioto2019rangenet}           & 29.2 && 65.9 & 00.9 & 00.0 & 00.0 & 84.4 & 31.1 & 01.1 & 45.6 & 40.9 & 68.5 & 37.0 & 88.5 & 37.3 & 53.8 \\
    RangenetSqueezeSegV2++~\cite{milioto2019rangenet} & 30.0 && 67.6 & 00.3 & 00.0 & 00.0 & 85.3 & 31.5 & 01.0 & 46.9 & 42.7 & 69.6 & 39.5 & 88.8 & 40.1 & 56.0 \\
    RangenetSqueezeSeg++~\cite{milioto2019rangenet}   & 30.6 && 72.2 & 00.7 & 00.0 & 00.0 & 84.7 & 35.9 & 01.1 & 47.9 & 43.7 & 74.7 & 38.6 & 88.9 & 37.4 & 56.5 \\
    PointTransformerV2~\cite{wu2022ptv2}              & 47.5 && 93.2 & 65.3 & 09.2 & 07.2 & 85.3 & 48.9 & 01.4 & 89.6 & 75.7 & 94.5 & 60.5 & 90.7 & 89.0 & 91.6 \\
    FRNet~\cite{xu2025frnet}                  & 53.3 && 86.7 & 13.6 & 00.1 & 05.5 & 87.6 & 48.5 & 14.8 & 83.5 & 70.5 & 94.4 & 57.7 & 90.9 & 74.8 & 71.7 \\
    \bottomrule
  \end{tabularx}%
  }
  \caption{
  Validation performance of different semantic segmentation methods on our custom dataset.
  We do not report scores for absent classes in the validation set, that includes: parking, motorcycle and motorcyclist, bicycle and bicyclist.
  }
  \label{tab:results}
\end{table*}
\begin{table}[t]
    \centering
    \begin{tabularx}{\linewidth}{l X YYY}
    \toprule
         Method && PQ & SQ & RQ \\
     \midrule
         Mask4Former-3D~\cite{yilmaz2024mask4former} && 42.5 & 61.3 & 52.8 \\
         Mask-PLS~\cite{marcuzzi2023maskpls} && 53.5 & 64.7 & 64.5 \\
         
    \bottomrule
    \end{tabularx}
    \caption{
    \textbf{Panoptic Segmentation Scores.}
    Two point cloud panoptic segmentation methods trained and evaluated on Panoptic-CUDAL.
    We observe acceptable performance on the validation dataset.
    }
    \label{tab:panotic_results}
\end{table}

\textbf{Labeling.}
A total of six sequences were recorded for annotation, 5 for training and 1 for validation.
Each sequence consists of about $2500$ point clouds and one sequence of about $2200$ point clouds with odometry poses, corresponding to the duration of 4 minutes 10 second sequences and 3 minutes 38 second sequences.
Annotated scenes provide detailed labels suitable for training and evaluation of LiDAR-based data processing methods.
We use the labeling scheme of the SemanticKITTI dataset~\cite{behley2019semantickitti,cordts2016cityscapes}, with the same class definitions.
Our dataset contains most of the labels present in the SemanticKITTI dataset, as shown in Figure~\ref{fig:labeldist}. 
We see an expected increase in the number of vegetation and terrain points, with fewer points for buildings or cars.
However, classes such as motorcyclist or motorcycle and bicycle or bicyclist are underrepresented due to the location of the recorded dataset.

\subsection{Data Annotation}
\begin{figure*}[!t]
  \centering
  \setlength{\tabcolsep}{2pt}
  \renewcommand{\arraystretch}{0.5}

  \begin{tabular}{ccc}
    \includegraphics[width=0.32\textwidth]{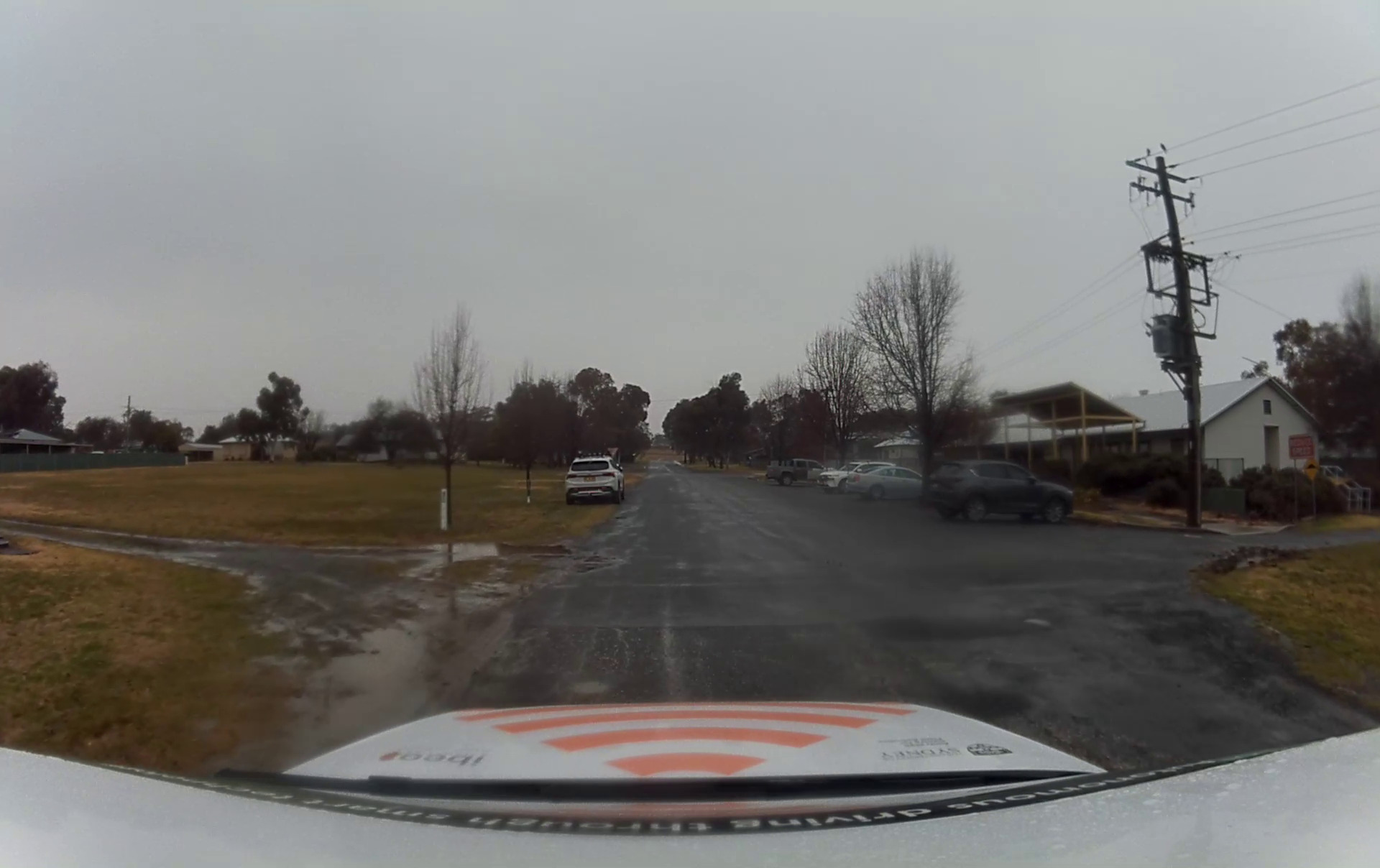} &
    \includegraphics[width=0.32\textwidth]{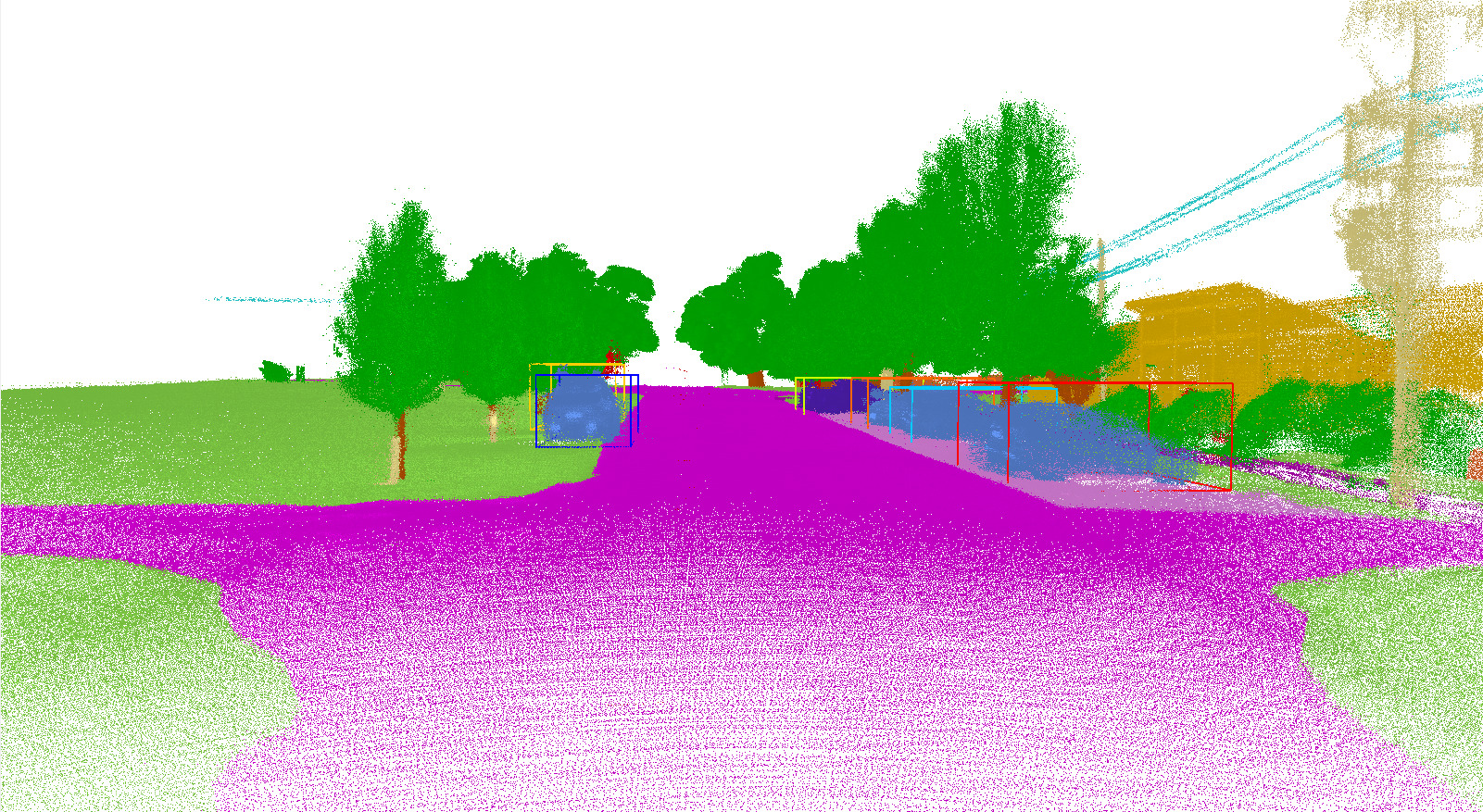} &
    \includegraphics[width=0.32\textwidth]{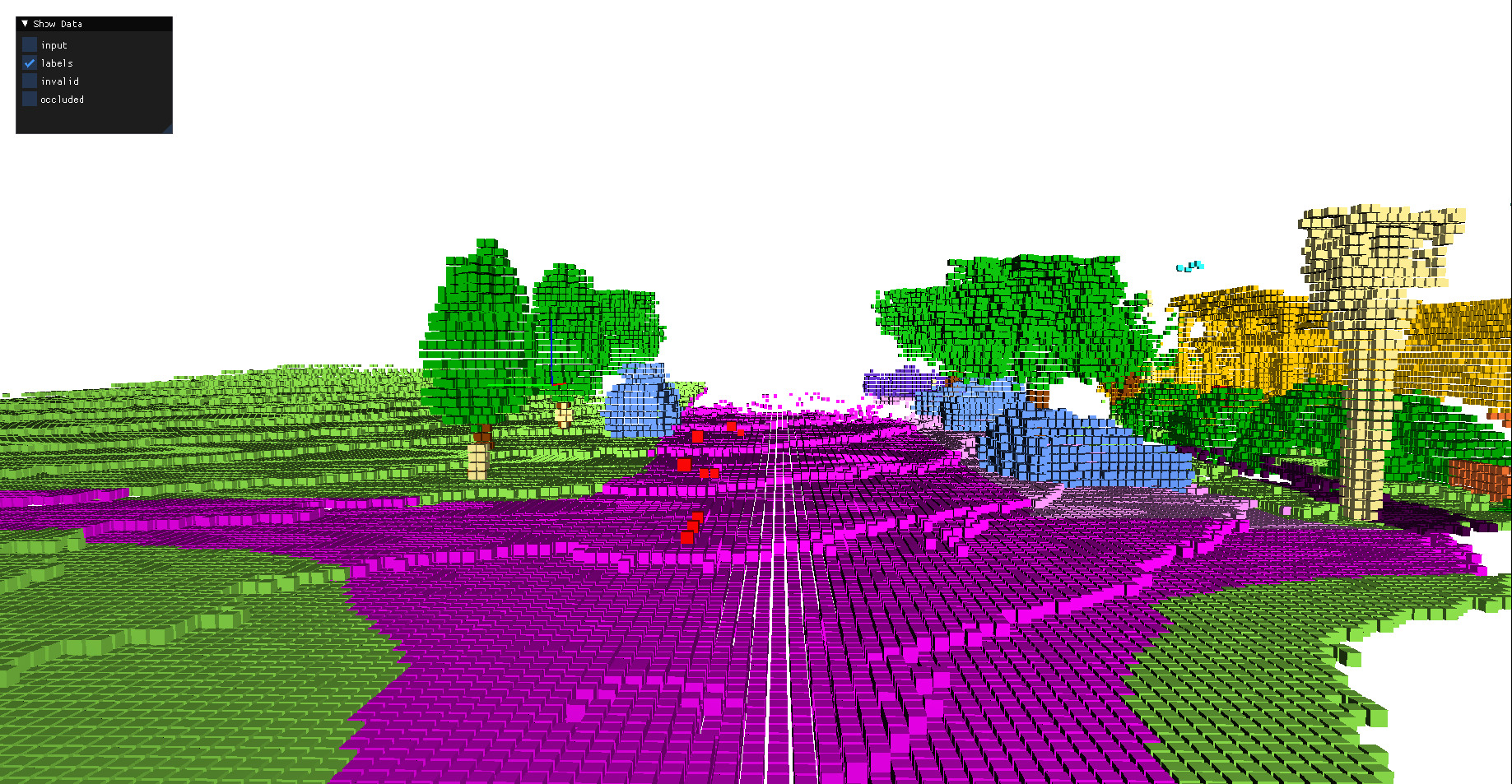} \\
    \includegraphics[width=0.32\textwidth]{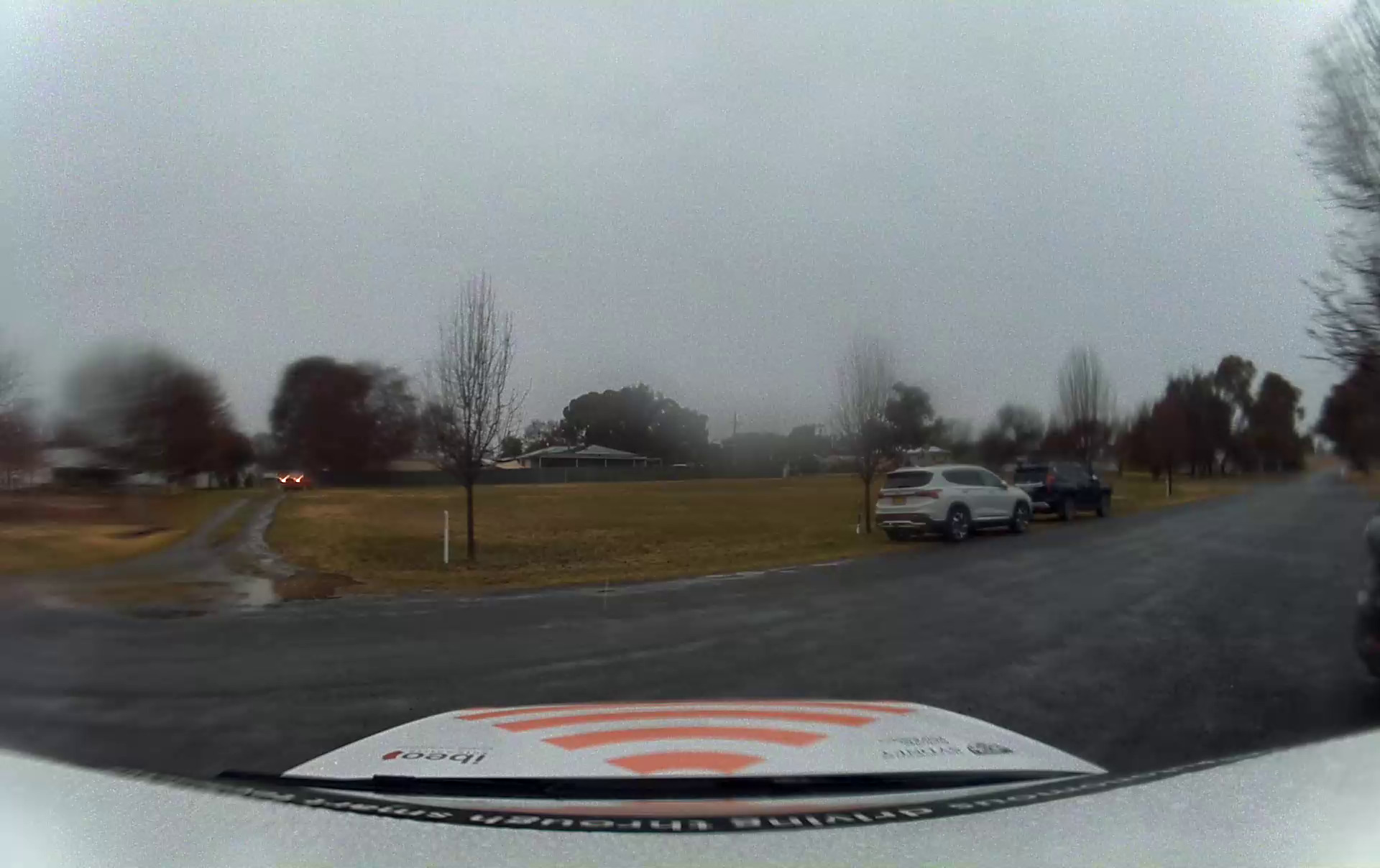} &
    \includegraphics[width=0.32\textwidth]{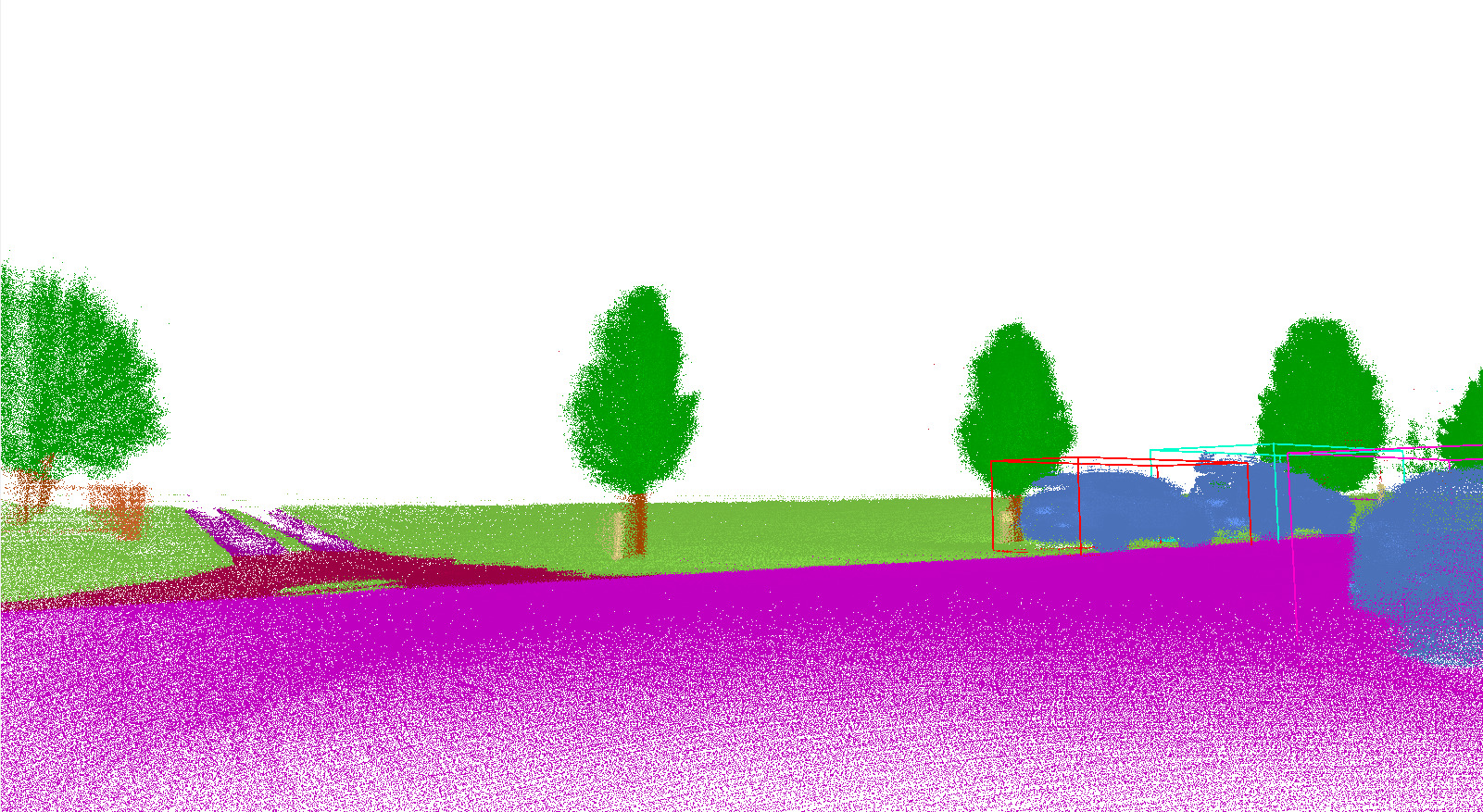} &
    \includegraphics[width=0.32\textwidth]{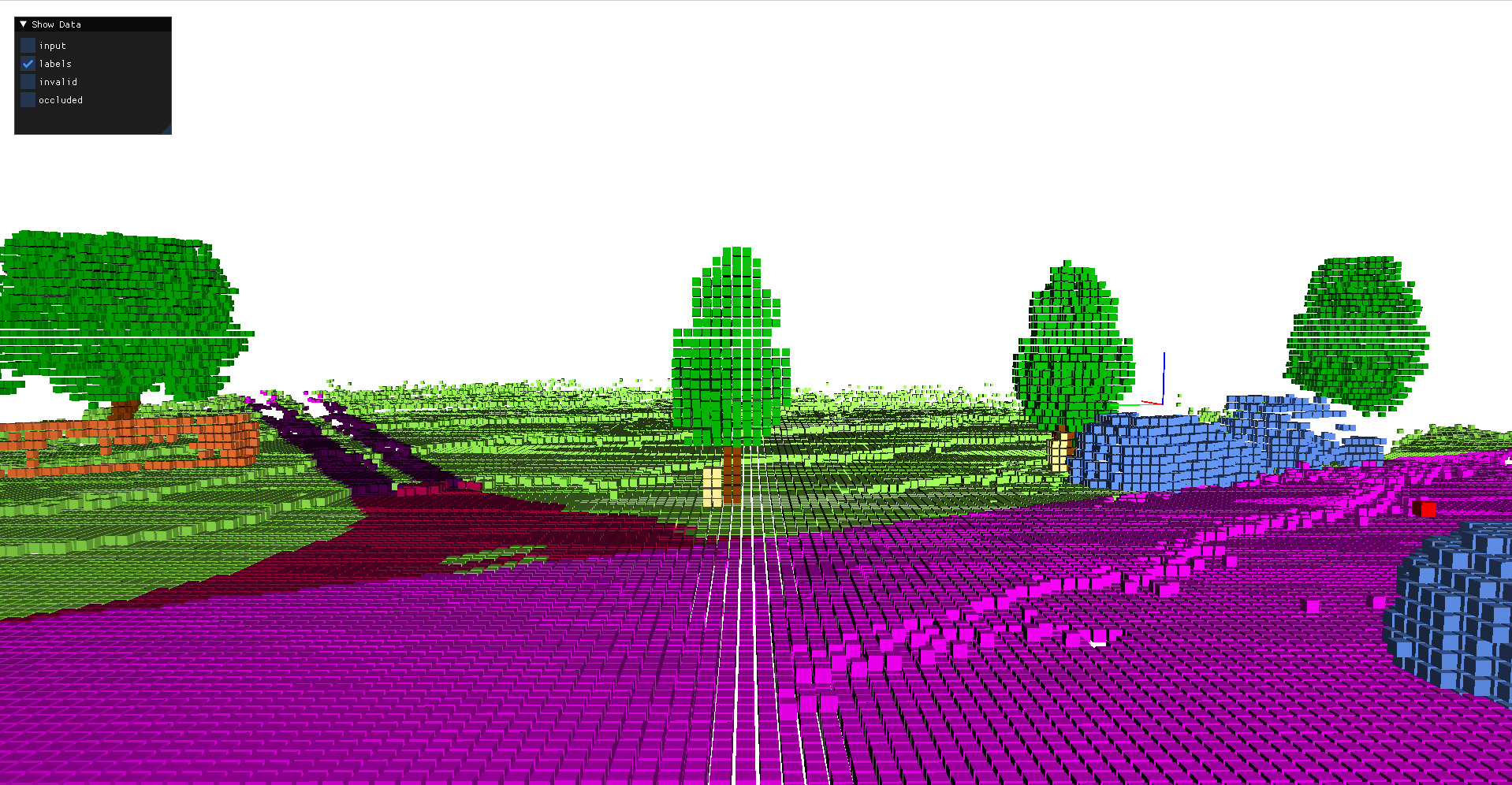} \\
    \includegraphics[width=0.32\textwidth]{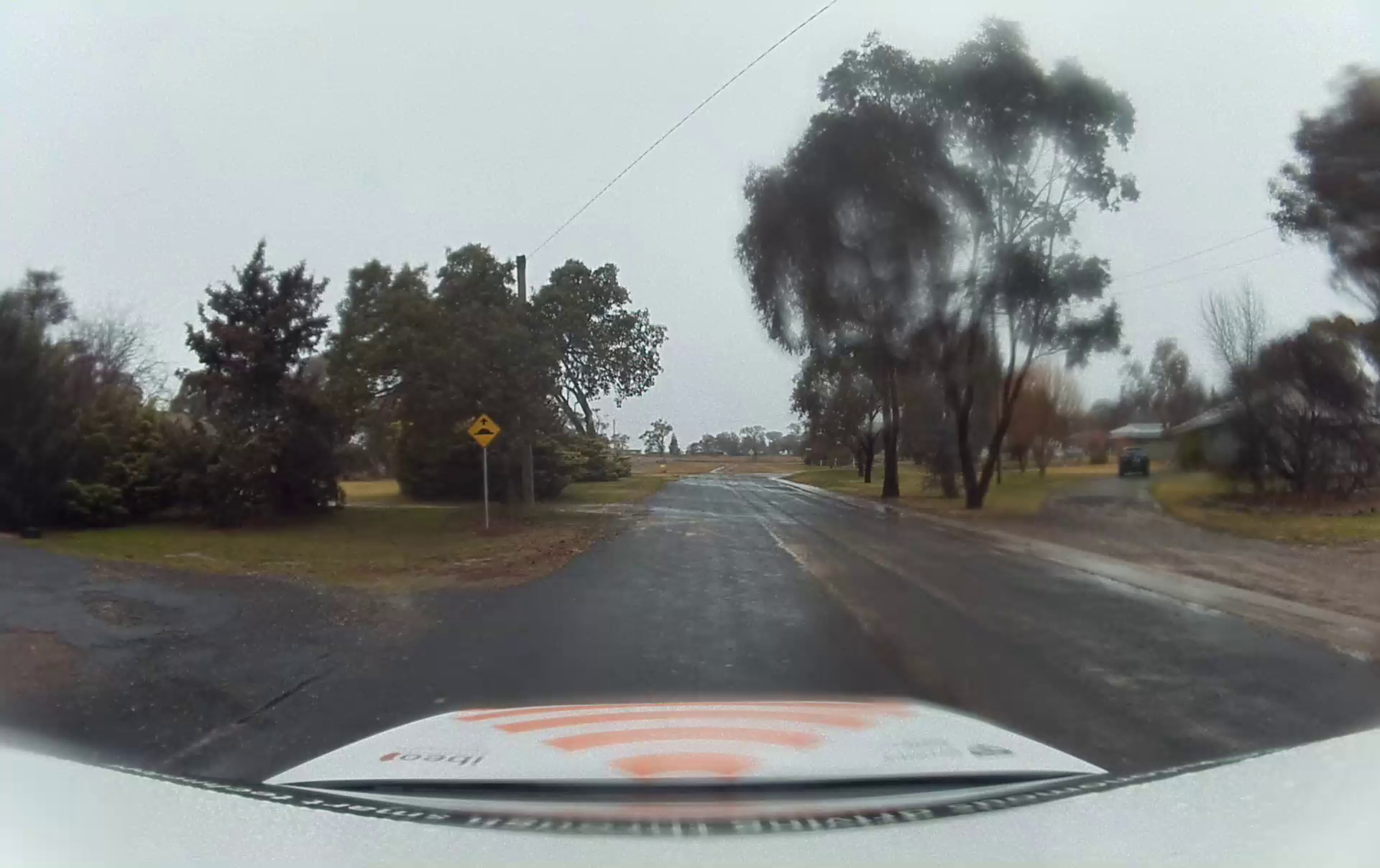} &
    \includegraphics[width=0.32\textwidth]{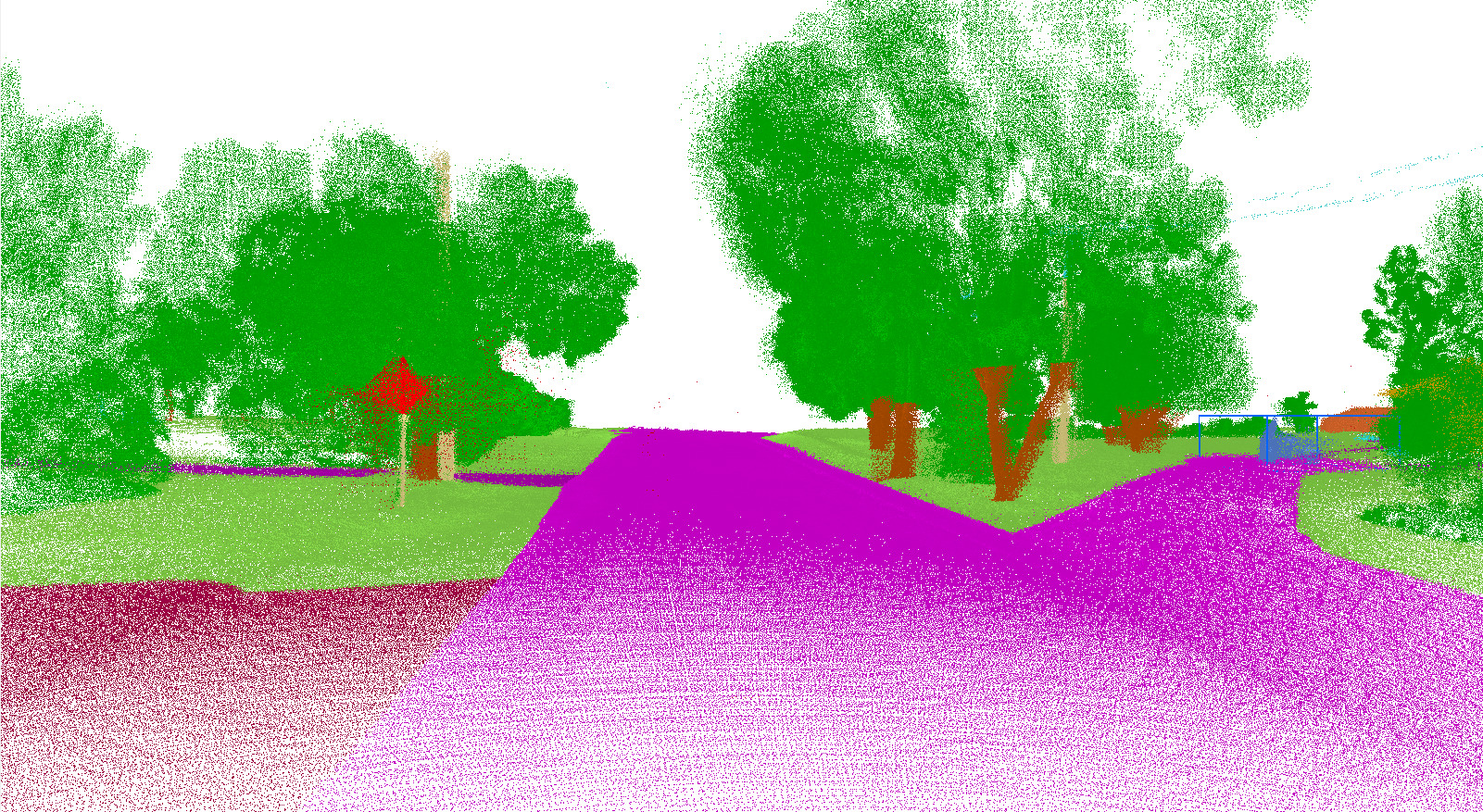} &
    \includegraphics[width=0.32\textwidth]{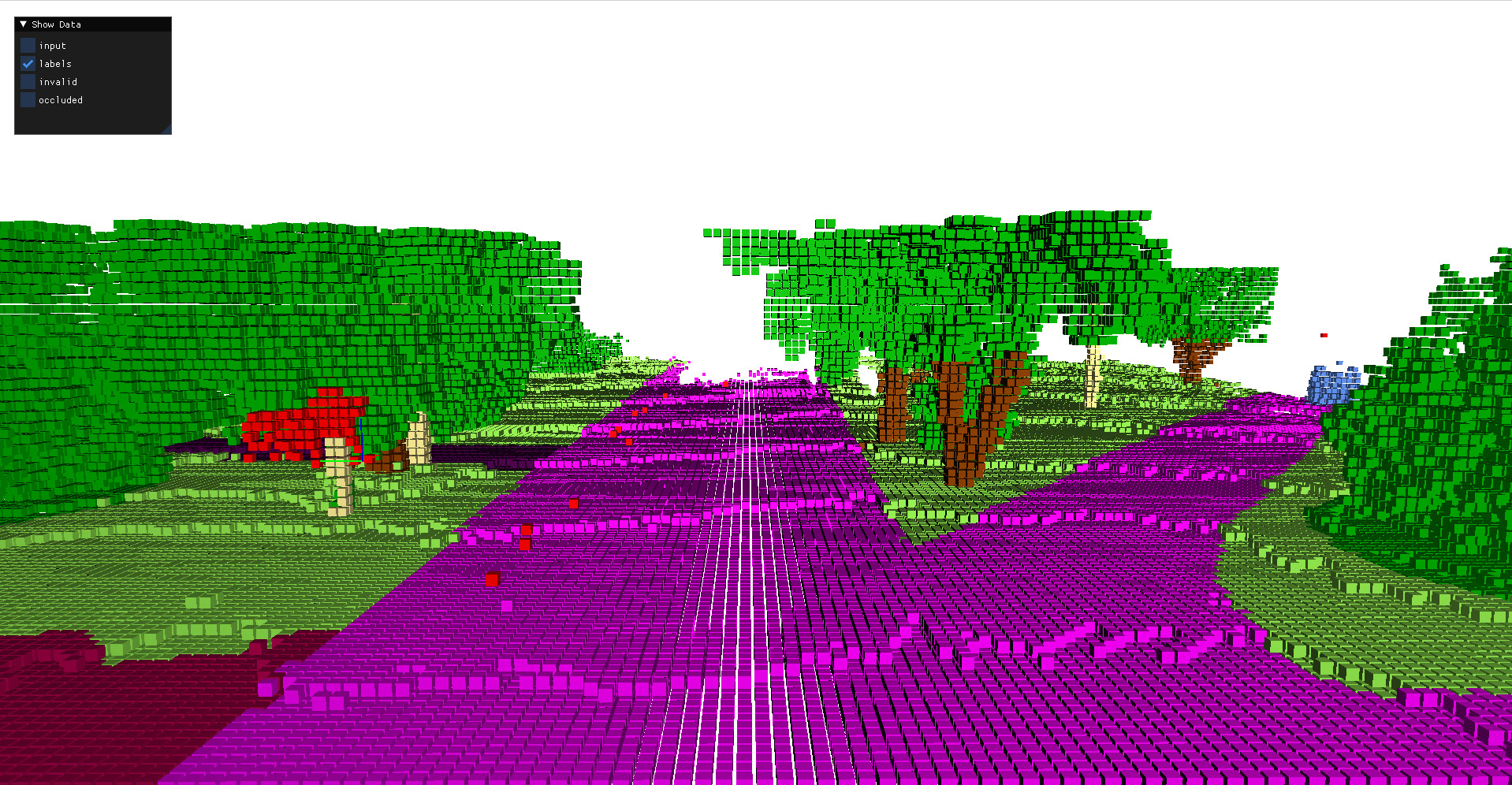} \\
    \includegraphics[width=0.32\textwidth]{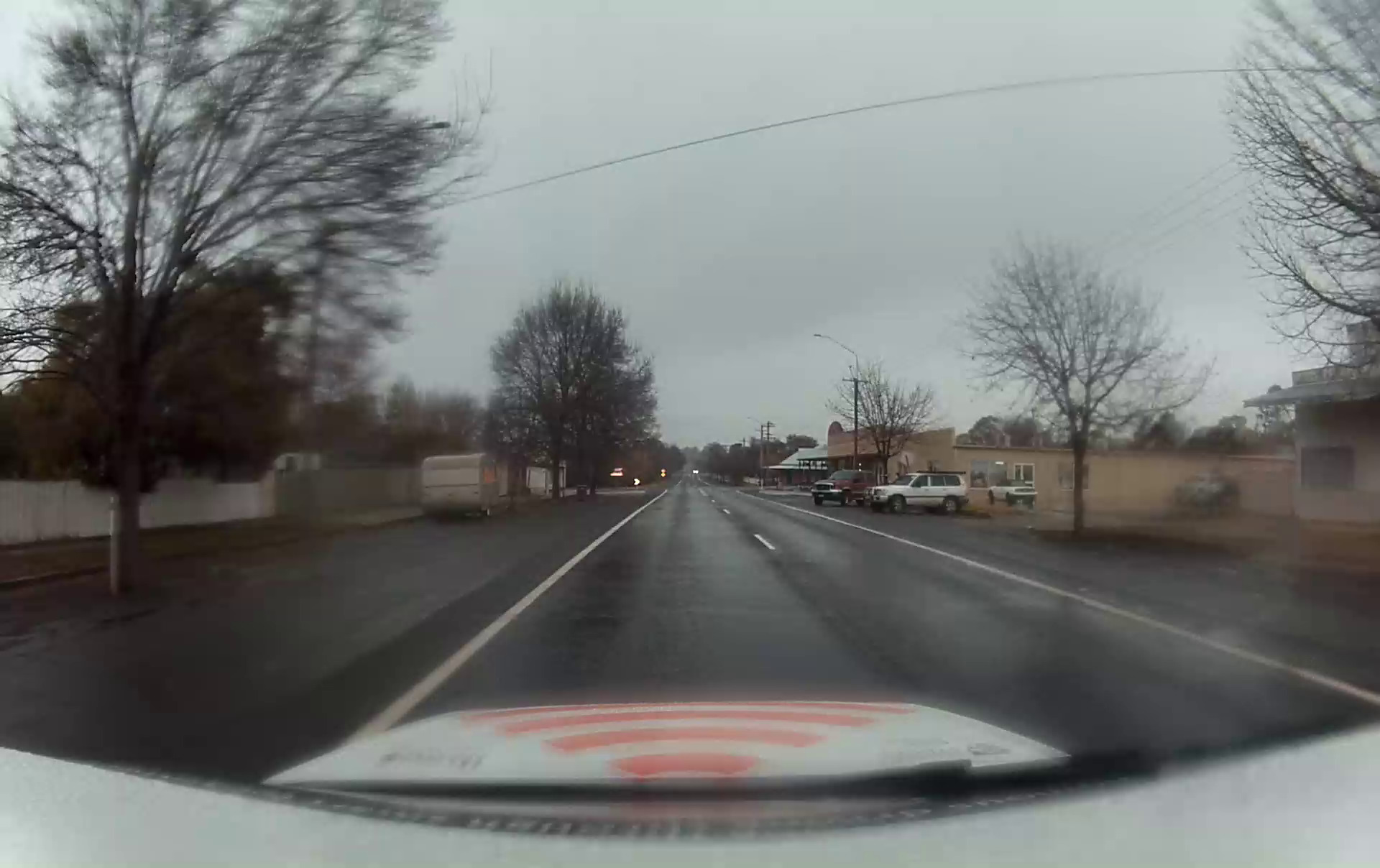} &
    \includegraphics[width=0.32\textwidth]{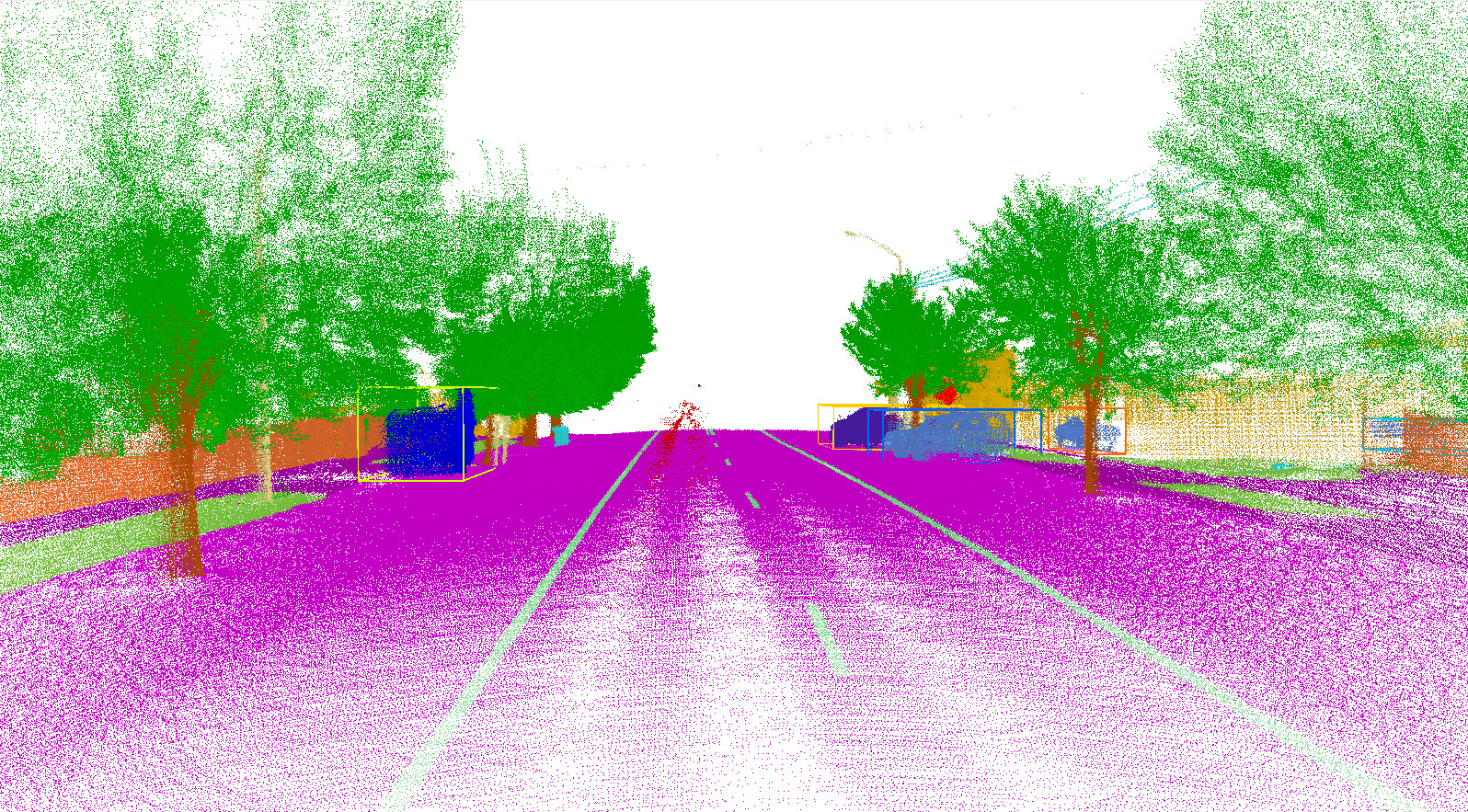} &
    \includegraphics[width=0.32\textwidth]{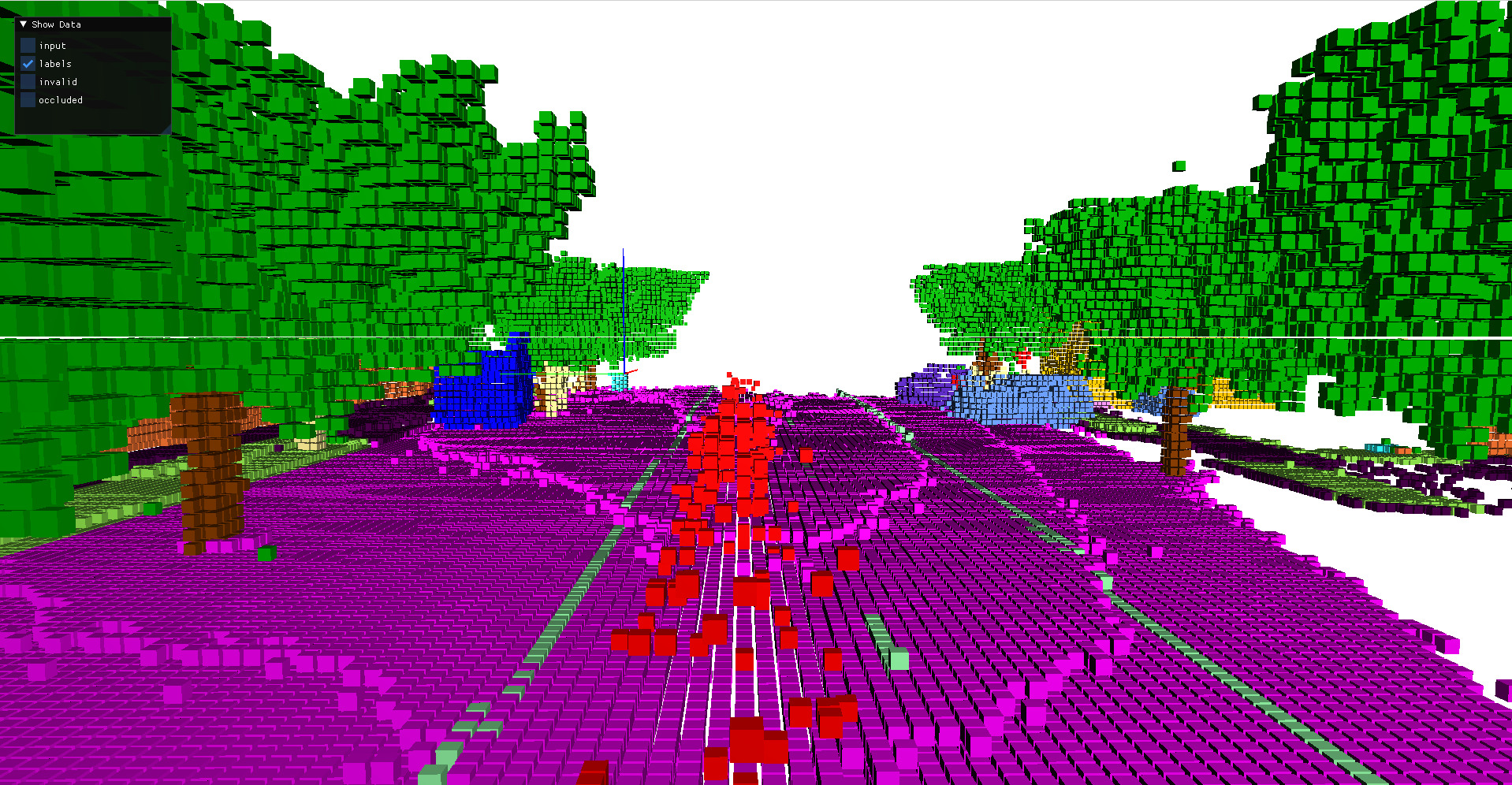}
  \end{tabular}

  \caption{Left: Raw image from our reference camera, with the resolution of 1920$\times$1208. Middle: Labeled point clouds with 20 semantic classes and instances. Right: Voxelized ground truth with voxel size of 0.2.}
  \label{fig:visual}
\end{figure*}

The data was annotated by four annotators working in parallel using the same annotation tool, as shown in Figure ~\ref{fig:tool}.
We annotated the dataset in four stages: semantic annotation, instance annotation, voxel annotation, and verification.
In the initial stage of annotation, two annotators gave semantic labels to point clouds, and for classes of objects that were not clearly visible, labels were verified using surround-view images.
The other two annotators then gave points belonging to each `thing' class an instance label consistent through the sequence. Then we use the voxelizer \cite{behley2019semantickitti} to generate the ground truth with the voxel size of 0.2, as shown to the right in ~\ref{fig:visual}.
In the last stage, the dataset was cross-checked by both of the annotation parties to verify consistency of the labels.

\section{Experiments and Results}
All experiments were trained all models using publicly available codebases with minimal changes. Each model is trained on five sequences and validated on one sequence. We initialize from SemanticKITTI-pretrained weights and fine-tune on Panoptic-CUDAL. This approach ensures consistency with existing benchmarks while adapting the models to our domain.
\subsection{Task and Metrics}
We focus on two tasks: \textbf{LiDAR semantic segmentation} and \textbf{3D semantic occupancy prediction}. For semantic segmentation,
the task assigns a class label to every point in a point cloud to enable detailed understanding of the scene.
For evaluation, we used the mean Intersection over Union (mIoU), which is a commonly used metric of semantic segmentation quality.
Since our dataset contains instance information, in addition to semantic segmentation methods, we train and evaluate panoptic segmentation methods.
To evaluate performance of panoptic segmentation, we use the Panoptic Quality (PQ) metric~\cite{fong2021panoptic}, that is a combination of segmentation quality (SQ) and recognition quality (RQ).

For 3D semantic occupancy prediction, we voxelize the 3D space surrounding the vehicle and assign semantic labels to each voxel based on camera or multi-modal sensor inputs. This task provides a unified modeling of static infrastructure and dynamic agents, critical for autonomous driving systems operating under different environmental conditions. Evaluation is performed using occupancy IoU for binary voxel occupancy prediction, and semantic mIoU across the 20 classes of SemanticKITTI~\cite{behley2019semantickitti}.

\subsection{Baseline Approaches}

\textbf{Semantic Segmentation.}
We use three types of models that use different point internal data representation.
We train projection-based RangeNet++~\cite{milioto2019rangenet} models, PointTransformerV2~\cite{wu2022ptv2} that directly works in 3D domain, and FRNet~\cite{xu2025frnet} that combines projection with direct point clouds processing.
Additionally, we train RangeNet++~\cite{milioto2019rangenet} method with different four different backbones.
The implementations were directly adopted from the public repositories~\cite{milioto2019rangenet,xu2025frnet,wu2022ptv2} and we follow training settings for the SemanticKITTI dataset~\cite{behley2019semantickitti}.

\textbf{Panoptic Segmentation.}
Similarly, the training configuration was not modified for the panoptic task.
We used two transformer-based methods for panoptic segmentation: MaskPLS~\cite{marcuzzi2023maskpls} and Mask4Former~\cite{yilmaz2024mask4former}.
We note that Mask4Former is originally a 4D segmentation method that we use for the 3D task.
We adopt it specifically due to the additional bounding box regression, that shows better performance for outdoor scenes, otherwise the model is similar to Mask3D~\cite{schult2023mask3d}, we denote this model as Mask4Former-3D.
For the training of MaskPLS, we followed the original setup and used SegContrast~\cite{nunes2022segcontrast} pre-trained MinkowskiNet~\cite{choy2019minkowski} backbone.
We train both models on our proposed dataset and follow SemanticKITTI training settings from original repositories.

\textbf{Occupancy Prediction.} Two baseline models are evaluated: InverseMatrixVT3D~\cite{ming2024inversematrixvt3d} and OccFusion~\cite{zhang2024occfusion}.  
\begin{table*}[t]
  \centering
  \resizebox{1.0\textwidth}{!}{
  \begin{tabular}{c|c|cc|ccccccccccccccc}
    \toprule
    Method & Input Modality & IoU & mIoU &
    \rotatebox{90}{road} & \rotatebox{90}{sidewalk} & \rotatebox{90}{other-ground} & \rotatebox{90}{building} & \rotatebox{90}{car} & \rotatebox{90}{truck}& \rotatebox{90}{other-vehicle} & \rotatebox{90}{vegetation} & \rotatebox{90}{trunk} & \rotatebox{90}{terrain} & \rotatebox{90}{person} & \rotatebox{90}{fence} & \rotatebox{90}{pole} & \rotatebox{90}{traffic-sign} \\
    \midrule 
    InverseMatrixVT3D ~\cite{ming2024inversematrixvt3d} & C & 57.84 & 17.60 & 55.03
     & 6.25 & 1.36 &  14.38 & 26.20 & 8.84 & 0.00 & 50.04 & 27.09 & 57.95 & 0.00 & 32.03 & 26.11 & 29.21  \\
    OCCFusion ~\cite{zhang2024occfusion} & C+L & 64.98 & 20.98 &
    53.41 & 5.53 & 0.72 & 37.86 & 25.91 & 12.3  & 0.00 & 59.83 & 30.41 & 61.08 & 0.00  & 35.99 & 38.23 & 37.30  \\ 
    \bottomrule
  \end{tabular}
  }
  \caption{\textbf{3D semantic occupancy prediction results on Panoptic-CUDAL dataset.} Modality notion: Camera (C), Camera+LiDAR (C+L). We do not report scores for absent classes in the validation set, that includes: parking, motorcycle and motorcyclist, bicycle and bicyclist.}
  \label{tab:occ_results}
  \vspace{-4mm}
\end{table*}
InverseMatrixVT3D is purely vision‐based.  
It uses a ResNet-101 DCN+FPN backbone to extract image features at scales $1/8$, $1/16$, and $1/32$. These are mapped to multi‐scale 3D voxel grids ($200\times200\times16$ down to $25\times25\times2$) via pre‐computed sparse projection matrices. The resulting BEV planes and volumetric grids are jointly refined by a lightweight global–local attention module.  

OccFusion builds on this by adding a VoxelNet LiDAR branch.  
It fuses LiDAR BEV and 3D feature volumes with the camera features at each scale via dynamic 2D/3D SENet blocks. It produces the same range of voxel resolutions.  

Both models are evaluated based on occupancy IoU and semantic mIoU and used directly from the public repositories following the settings for the SemanticKITTI.
\vspace{-4mm}
\subsection{Experimental Results}
Table~\ref{tab:results} summarizes the semantic segmentation validation performance for each method. Transformer-based and hybrid models, such as PointTransformerV2 and FRNet, outperform projection-based approaches, particularly for common 'stuff' categories such as vegetation and terrain. In contrast, rare or visually ambiguous categories such as 'other ground' and 'other vehicle' consistently achieve lower intersection over union (IoU) scores, reflecting both class imbalance and the perceptual challenges. As the validation set lacks examples of 'motorcycle', 'motorcyclist', 'bicycle', 'bicyclist', and 'parking', these categories are excluded.

In Table~\ref{tab:panotic_results}, we report scores for the panoptic methods, which achieve acceptable results for the proposed dataset.
We observe that MaskPLS method achieves higher results compared to Mask4Former-3D. However, it is important to note that we use a pre-trained backbone for the model initialization.

Table~\ref{tab:occ_results} summarizes the results of the occupancy prediction validation. Both models are evaluated using the Scene Completion (SC) IoU and Semantic Scene Completion (SSC) mIoU metrics. We build on the pre-trained weights provided by~\cite{zhang2024occfusion}. Overall, the models perform comparably to results from other outdoor benchmarks for classes such as `vegetation', `terrain', and `buildings', which are segmented with high accuracy. However, the categories `other-vehicle' and `other-ground' appear sparsely in the training split.

The results in Table~\ref{tab:occ_results} also indicate that adding LiDAR to the camera stream increased SC IoU from 57.84\% to 64.98\% and SSC mIoU from 17.60\% to 20.98\%. However, on a wet country road, the additional depth data provided almost no benefit for certain classes: the IoU for `other-ground' dropped from 1.36\% to 0.72\% and the IoU for `pavement' dropped from 6.25\% to 5.53\%. These show that on wet, low-contrast surfaces and feature-poor open roads, LiDAR returns contribute little new semantic information.

\begin{figure}[t]
    \centering
    \begin{subfigure}[]{0.99\columnwidth}
        \centering        
        \includegraphics[trim={0cm 5cm 0 0},clip,width=\columnwidth]{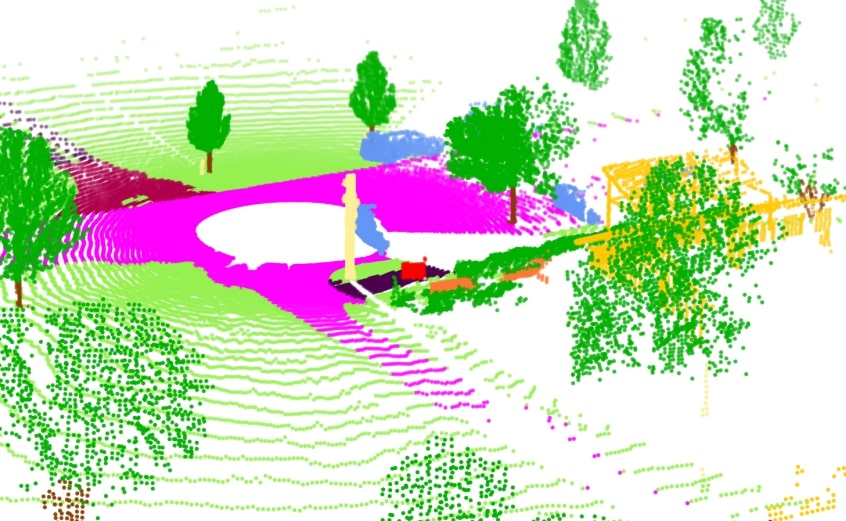}
        \caption{Ground Truth}
        \label{fig:gt}
    \end{subfigure}
    
    \begin{subfigure}[]{0.99\columnwidth}
        \centering
        \includegraphics[trim={0cm 5cm 0 0},clip,width=\columnwidth]{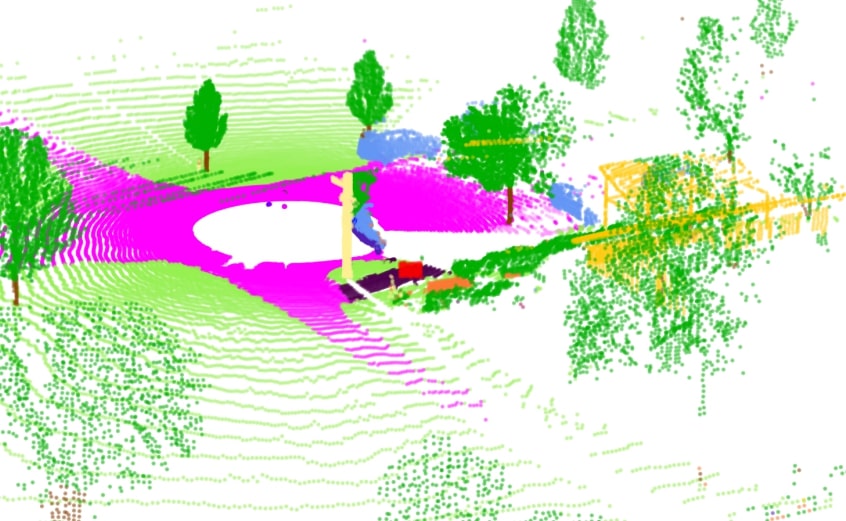}
        \caption{FRNet~\cite{xu2025frnet} Prediction}
        \label{fig:pred}
    \end{subfigure}
    \caption{Ground Truth vs Prediction in rural environments.}
    \label{fig:inference}
    \vspace{-4mm}
\end{figure}

The inference results in Fig.~\ref{fig:inference}
 show that fine-tuning FRNet~\cite{xu2025frnet} on Panoptic-Cudal yields sufficient accuracy for use in this environment.

\section{Conclusion}
This paper presents an outdoor dataset featuring high-resolution LiDAR, surround-view cameras and precise pose information, with panoptic labels for LiDAR point-cloud segmentation. The dataset captures several driving sequences in rainy conditions in a rural Australian area, addressing the lack of such data in the autonomous driving community. We analysed the label distribution and set baselines for semantic segmentation, panoptic segmentation and 3D occupancy prediction.
Although Panoptic-CUDAL offers extensive coverage of challenging rural scenarios, our findings reveal several areas that warrant further research. 
Due to the nature of rural settings, certain road users, such as cyclists and motorcyclists, are rarely seen. While this reflects real-world distributions, it poses challenges for training and evaluating models. Our baseline results demonstrate the limitations of applying models trained on urban, clear-weather datasets.


\PAR{Acknowledgments.}{ \small
A. Nekrasov and M. Burdorf acknowledge funding by BMBF project ``WestAI'' (grant no. 01IS22094D).
}

\bibliographystyle{IEEEtran}
\bibliography{main}

\end{document}